\documentclass[10pt,journal,compsoc]{IEEEtran}

\ifCLASSOPTIONcompsoc
  \usepackage[nocompress]{cite}
\else
  \usepackage{cite}
\fi
\ifCLASSINFOpdf
   \usepackage[pdftex]{graphicx}
   \usepackage{xcolor,tikz, subcaption, color}
\else
\fi
\usepackage{amsmath, amssymb, amsfonts, bm}

\usepackage[ruled, vlined, commentsnumbered, linesnumbered]{algorithm2e}
\usepackage{algorithmic}

\usepackage{booktabs,tabularx}
\usepackage[symbol]{footmisc}

\usepackage{array, multirow}

\usepackage{url, multirow, hyperref}
\usepackage{kotex}

\newcommand{\etal}{{\em et al.}}       
\newcommand{\eg}{{\em e.g.}}           
\newcommand{\ie}{{\em i.e.}}           

\newcommand{\ks}[1]{\textcolor{black}{#1}}

\newcommand{\tx}{\tilde{\mathbf{X}}}
\newcommand{\x}{\mathbf{X}}
\newcommand{\y}{\mathbf{y}}
\newcommand{\tar}{\mathbf{t}}
\newcommand{\M}{\mathbf{M}}

\newcommand{\HI}{\color{black}}
\newcommand{\cmt}{\color{black}}

\newcommand{\supp}{\color{black}}

\begin{document}
\title{Learn-Explain-Reinforce: Counterfactual Reasoning and Its Guidance to Reinforce an Alzheimer's Disease Diagnosis Model}

\author{Kwanseok~Oh$^{*}$,
        Jee~Seok~Yoon$^{*}$,
        and~Heung-Il~Suk,~\IEEEmembership{Senior Member,~IEEE}
        
\IEEEcompsocitemizethanks{
\IEEEcompsocthanksitem K. Oh is with the Department of Artificial Intelligence, Korea University, Seoul 02841, Republic of Korea (e-mail: ksohh@korea.ac.kr).
\IEEEcompsocthanksitem J. Yoon is with the Department of Brain and Cognitive Engineering, Korea University, Seoul 02841, Republic of Korea (e-mail: wltjr1007@korea.ac.kr).
\IEEEcompsocthanksitem H.-I. Suk is with the Department of Artificial Intelligence and the Department of Brain and Cognitive Engineering, Korea University, Seoul 02841, Republic of Korea and the corresponding author (e-mail: hisuk@korea.ac.kr).
\IEEEcompsocthanksitem $^{*}$ indicates equal contribution.
}
}

\markboth{Preprint}%
{Oh \MakeLowercase{\textit{et al.}}: \MakeUppercase{Learn-Explain-Reinforce: Counterfactual Reasoning and Its Guidance to Reinforce an Alzheimer's Disease Diagnosis Model}}

\IEEEtitleabstractindextext{%
\begin{abstract}
    Existing studies on disease diagnostic models focus either on diagnostic model learning for performance improvement or on the visual explanation of a trained diagnostic model. We propose a novel learn-explain-reinforce (LEAR) framework that unifies diagnostic model learning, visual explanation generation (explanation unit), and trained diagnostic model reinforcement (reinforcement unit) guided by the visual explanation. For the visual explanation, we generate a counterfactual map that transforms an input sample to be identified as an intended target label. For example, a counterfactual map can localize hypothetical abnormalities within a normal brain image that may cause it to be diagnosed with Alzheimer's disease (AD). We believe that the generated counterfactual maps represent data-driven and model-induced knowledge about a target task, \ie, AD diagnosis using structural MRI, which can be a vital source of information to reinforce the generalization of the trained diagnostic model. To this end, we devise an attention-based feature refinement module with the guidance of the counterfactual maps. The explanation and reinforcement units are reciprocal and can be operated iteratively. Our proposed approach was validated via qualitative and quantitative analysis on the ADNI dataset. Its comprehensibility and fidelity were demonstrated through ablation studies and comparisons with existing methods.
\end{abstract} 

}

\maketitle

\IEEEdisplaynontitleabstractindextext

\IEEEpeerreviewmaketitle

\IEEEraisesectionheading{\section{Introduction}\label{sec:introduction}}

\IEEEPARstart{A}{lzheimer's} disease (AD) is known as one of the most prevalent neurodegenerative diseases, characterized by progressive and irreversible memory loss and cognitive function decline or impairment~\cite{alzheimer20192019}. AD causes the damage and destruction of nerve cells in brain regions related to memory, language, and other cognitive functions, and it has contributed to 60--80$\%$ of the world's dementia cases~\cite{alzheimer20182018}.
Brain atrophy associated with AD emerges as a {\em continuous} progression from cognitively normal (CN) to mild cognitive impairment (MCI) and dementia in the symptomatic spectrum~\cite{li2013variation}. Currently available AD-related medicines have marginal effect in alleviating amnesic symptoms or slowing their progression. Thus, early detection and timely intervention of AD at its preclinical or prodromal stages are of paramount importance in the prevention of its progression and in diminishing its incidence.

Of various brain imaging tools, structural magnetic resonance imaging (sMRI) has been most intensively studied for AD diagnosis as it provides imaging biomarkers of neuronal loss in the anatomical structures of a brain~\cite{frisoni2010clinical}. Specifically, sMRI scans are helpful in detecting and measuring morphological changes in the brain, such as enlarged ventricle and regional atrophies, and anatomical variations across subjects.
In the last few decades, researchers have devoted their efforts to devising machine-learning techniques that can analyze and identify the potential risk of a subject having AD or MCI at an early stage~\cite{kloppel2008automatic,hinrichs2009spatially,zhang2011multimodal}. More recently, with the unprecedented advances in deep learning, there have been many successful studies in sMRI-based AD diagnosis that achieved clinically applicable performance~\cite{suk2013deep,korolev2017residual,suk2017deep}.

In the meantime, there has been a growing need for explainability of a model's output and/or interpretability of a model's internal workings~\cite{arrieta2020explainable,singh2020explainable}. The black-box nature of deep learning models limits their real-world application in the fields of medicine, security, and finance, especially where fairness, accountability, and transparency are essential. From the end-user's (\eg, clinicians and patients) point of view, it is crucial to be able to interpret and explain a deep-learning model's output at the level of human knowledge and understanding. However, building a predictive model for high performance that is also equipped with interpretability or explainability is still an unsolved problem because of their trade-off, \ie, interpretable/explainable models tend to have lower performance than black-box models~\cite{gunning2019xai,mascharka2018transparency,rudin2019stop}, especially in the field of medical vision.

Reducing this trade-off between performance and interpretability/explainability has been a long-standing goal in the field of explainable AI (XAI). In the early era of XAI~\cite{gilpin2018explaining}, researchers have proposed various methods for discovering or identifying the regions that have the most influence on deriving the outcome of a classifier~\cite{selvaraju2017grad,sundararajan2017axiomatic,bach2015pixel,montavon2017explaining,shrikumar2017learning,zeiler2014visualizing}. The main objective of those XAI methods is to answer the question, {``\em For an input $X$, which part influenced the classifier's decision to label it $Y$?''} However, recent XAI methods try to answer the question that can offer a more fundamental explanation: {``\em If an input $X$ was $X^*$, would the outcome have been $Z$ rather than $Y$?''}~\cite{goyal2019counterfactual,goyal2019explaining,wang2020scout} in the sense of causality. This sort of explanation is defined at the root of {\em counterfactual reasoning}~\cite{wachter2017counterfactual}. Counterfactual reasoning can provide an explanation at the level of human knowledge as it explains a model’s decision in hypothetical scenarios.

Inspired by this philosophical concept of counterfactual reasoning, in this work, we propose a novel method for a higher-level visual explanation of a deep predictive model designed and trained for AD diagnosis using sMRI. 
Specifically, our method generates a `\emph{counterfactual map}' conditioned on a target label (\ie, hypothetical scenario). This map is added to the input image to transform it to be diagnosed as a target label.
For example, when a counterfactual map
is added to the input MRI of AD subject, it causes the input MRI to be transformed such that it will be diagnosed as CN~\cite{baumgartner2018visual,bass2020icam}.
%
Most of the existing works on producing a counterfactual explanation exploit generative models with generative adversarial network (GAN) and its variants~\cite{chang2018explaining,van2019interpretable,sauer2021counterfactual}.
To the best of our knowledge, however, they are limited to producing a single-way~\cite{dash2020counterfactual,goyal2019explaining,chang2018explaining} or dual-way~\cite{goyal2019counterfactual,wang2020scout} explanation. In other words, they only consider one or two hypothetical scenarios for counterfactual reasoning (\eg, single-way counterfactual map can only transform a CN subject to an AD patient, and vice versa for dual-way maps).
Thus, when there are more than two classes of interest for diagnosis, \eg, CN \textit{vs.} MCI \textit{vs.} AD, a set of such explainable models must be built separately and independently for different pairs of clinical labels, \eg, CN \textit{vs.} MCI, MCI \textit{vs.} AD, and CN \textit{vs.} AD. However, with those separately and independently trained explanation models, there is likely to be incompatibility and inconsistency in explanation, especially, in terms of the AD spectrum, raising accountability or interpretability issues. Consequently, it is necessary to build a single model for multi-way counterfactual map generation.
Notably, a multi-way counterfactual map for an AD diagnostic model can provide a natural proxy for the stratification of diseases by producing hypothetical scenarios for intermediate stages (\eg, CN→MCI→AD) of a disease. To this end, we propose a novel \textit{multi-way} counterfactual reasoning method such that we can produce counterfactual maps for transforming an input to be any of the clinical labels under consideration (\ie, CN, MCI, and AD).

\begin{figure}[t]
    \centering
    \includegraphics[width=.9\linewidth]{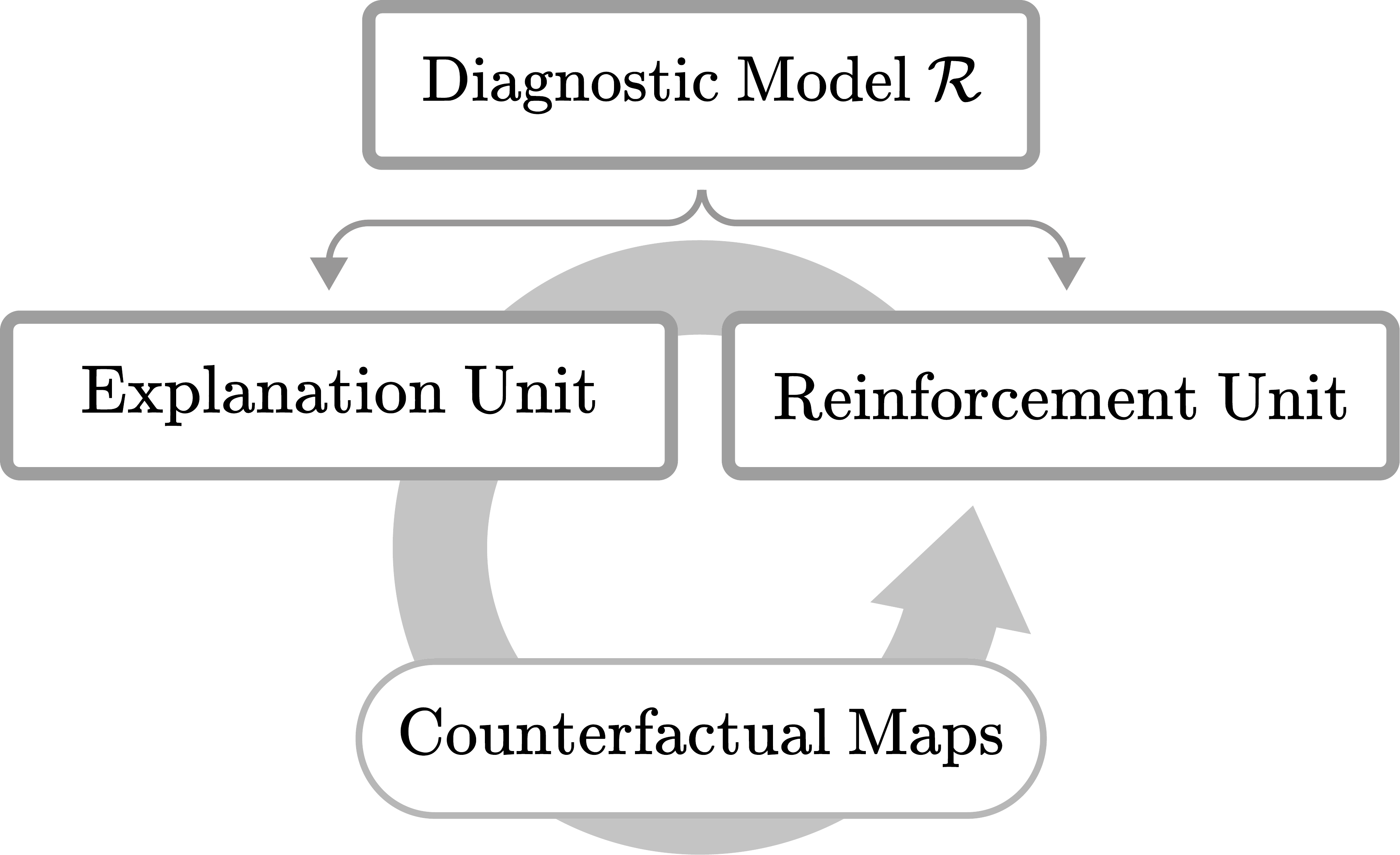}
    \caption[]{Schematic of our proposed learn-explain-reinforce (LEAR)\footnotemark[2]~framework. The explanation unit is a variation of conditional GAN that can synthesize a counterfactual map conditioned on an arbitrary target label. The reinforcement unit provides adequate guidance from the produced counterfactual map for reinforcing the performance of diagnostic model $\mathcal{R}$. We also introduce a simple iterative optimization scheme that enables simultaneous improvement of the explanation and diagnostic performance.
    }\label{fig:overview}
\end{figure}

Meanwhile, we believe it is desirable to utilize the counterfactual maps as \emph{privileged information}, derived from an explanation model in combination with an AD diagnostic model during a training stage, to further enhance a classifier's generalizability, thus improving performance. In particular, thanks to the favorable counterfactual map's localization properties, we propose to exploit such information to guide a diagnostic model's focus on learning representations and discovering disease-related discriminative regions, which can be regarded as anatomical landmarks for diagnosis.
\footnotetext[2]{The term reinforce/reinforcement used in this article refer to reinforcing the visual explanation and diagnostic models by means of attention-based guidance. Thus, it should \textit{not} be confused with reinforcement learning in machine learning, which is a learning paradigm for intelligent agents.}

To this end, we propose a novel learn-explain-reinforce (LEAR)\footnotemark[2] framework.
Our LEAR framework can produce high-quality counterfactual maps with state-of-the-art diagnostic performances through explanation-guided model reinforcement. {\HI Fig.~\ref{fig:overview} illustrates the schematic diagram of our proposed framework for counterfactual map generation to explain a diagnostic model's output (Explanation Unit) and its use to reinforce the generalizability of the diagnostic model via our newly devised pluggable reinforcement unit.}

\begin{table*}[t]\scriptsize\setlength{\tabcolsep}{5.pt}
    \caption{Recent studies categorized into visual explanation, use of attention mechanism, explanation-guided methods, and ability to reinforce visual explanation.}
    \label{tab:rel_works}
    \centering
    \begin{tabular}{cccccc}
    \toprule
    \multicolumn{1}{c}{Methods} & \multicolumn{1}{c}{Visual Explanation} & \multicolumn{1}{c}{Attention} & \multicolumn{1}{c}{Guidance} & \multicolumn{1}{c}{Reinforce} & \multicolumn{1}{c}{Description}\\
\toprule
    \multicolumn{1}{c}{\multirow{2}{*}{Liu \etal~\cite{liu2020design}}} & \multicolumn{1}{c}{\multirow{2}{*}{LRP~\cite{bach2015pixel}}} & \multicolumn{1}{c}{\multirow{2}{*}{}} &\multicolumn{1}{c}{\multirow{2}{*}{}} &\multicolumn{1}{c}{\multirow{2}{*}{}} &\multicolumn{1}{c}{Improving the diagnostic performance through instance normalization}\\
    &&&&&\multicolumn{1}{c}{\text{and model capacity increase}}\\
\midrule
    \multicolumn{1}{c}{\multirow{2}{*}{Korolev \etal~\cite{korolev2017residual}}} & \multicolumn{1}{c}{\multirow{2}{*}{-}} & \multicolumn{1}{c}{\multirow{2}{*}{}} &\multicolumn{1}{c}{\multirow{2}{*}{}} &\multicolumn{1}{c}{\multirow{2}{*}{}} &\multicolumn{1}{c}{Unique feature extraction by applying the dropout operation before}\\
    &&&&&\multicolumn{1}{c}{\text{the fully connected layer}}\\
\midrule
    \multicolumn{1}{c}{\multirow{2}{*}{Jin \etal~\cite{jin2019attention}}} & \multicolumn{1}{c}{\multirow{2}{*}{-}} & \multicolumn{1}{c}{\multirow{2}{*}{$\checkmark$}} &\multicolumn{1}{c}{\multirow{2}{*}{}} &\multicolumn{1}{c}{\multirow{2}{*}{}} &\multicolumn{1}{c}{Discriminative feature extraction using the attention-based}\\
    &&&&&\multicolumn{1}{c}{\text{residual network}}\\
\midrule
    \multicolumn{1}{c}{\multirow{2}{*}{Zhang \etal~\cite{zhang2021explainable}}} & \multicolumn{1}{c}{\multirow{2}{*}{Grad-CAM~\cite{selvaraju2017grad}}} & \multicolumn{1}{c}{\multirow{2}{*}{$\checkmark$}} &\multicolumn{1}{c}{\multirow{2}{*}{}} &\multicolumn{1}{c}{\multirow{2}{*}{}} &\multicolumn{1}{c}{Global and local representation captured using self-attention with}\\
    &&&&&\multicolumn{1}{c}{\text{the residual connection}}\\
\midrule
    \multicolumn{1}{c}{\multirow{2}{*}{Lian \etal~\cite{lian2020attention}}} & \multicolumn{1}{c}{\multirow{2}{*}{CAM~\cite{zhou2016learning}}} & \multicolumn{1}{c}{\multirow{2}{*}{$\checkmark$}} &\multicolumn{1}{c}{\multirow{2}{*}{$\checkmark$}} &\multicolumn{1}{c}{\multirow{2}{*}{}}&\multicolumn{1}{c}{Attention-guided anatomical landmarks to capture multi-level}\\
    &&&&&\multicolumn{1}{c}{\text{discriminative patches and regions}}\\
\midrule
    \multicolumn{1}{c}{\multirow{2}{*}{Li \etal~\cite{li2019novel}}} & \multicolumn{1}{c}{\multirow{2}{*}{CAM~\cite{zhou2016learning}}} & \multicolumn{1}{c}{\multirow{2}{*}{}} &\multicolumn{1}{c}{\multirow{2}{*}{$\checkmark$}} &\multicolumn{1}{c}{\multirow{2}{*}{$\checkmark$}} &\multicolumn{1}{c}{Iterative attention-focusing strategy for joint pathological region}\\
    &&&&&\multicolumn{1}{c}{\text{localization and identification}}\\
\midrule
    \multicolumn{1}{c}{\multirow{2}{*}{\textbf{LEAR (Ours)}}} & \multicolumn{1}{c}{\text{Counterfactual}} & \multicolumn{1}{c}{\multirow{2}{*}{$\checkmark$}} &\multicolumn{1}{c}{\multirow{2}{*}{$\checkmark$}} &\multicolumn{1}{c}{\multirow{2}{*}{$\checkmark$}} &\multicolumn{1}{c}{\text{Reinforcement of the diagnostic performance and explainability via}} \\
    &\multicolumn{1}{c}{\text{Reasoning}~\cite{wachter2017counterfactual}}&&&&\multicolumn{1}{c}{\text{the self-iterative training strategy with guidance}}\\
\bottomrule
\end{tabular}
\end{table*}

The main contributions of our work can be summarized as follows:
    \begin{itemize}
        \item {We propose a novel learn-explain-reinforce framework that integrates the following tasks: (1) training a diagnostic model, (2) explaining a diagnostic model's output, and (3) reinforcing the diagnostic model based on the explanation systematically. To the best of our knowledge, this work is the first that exploits an explanation output to improve the generalization of a diagnostic model reciprocally.}
        \item {In regard to explanation, we propose a GAN-based method to produce \emph{multi-way} counterfactual maps that can provide a more precise explanation, accounting for severity and/or progression of AD.}
        \item {Our work qualitatively and quantitatively surpasses state-of-the-art works in visual explanation and classification performance simultaneously.}
    \end{itemize}

{\cmt The remainder of this article is organized as follows. In Section~\ref{sec2}, we briefly review related work on attribution-based approaches and counterfactual explanations. Next, we introduce the automated AD diagnosis using attention with guidance. The proposed method is described in detail in Section~\ref{sec3}. In Section~\ref{sec4}, we describe the studied datasets (\ie, ADNI-1 and ADNI-2) with the data preprocessing pipeline as well as the experimental settings, competing methods, and qualitative and quantitative experimental results. We conclude this article and briefly discuss our stance on model explanation in Section~\ref{sec5}.} Our code is available at: \url{https://github.com/ku-milab/LEAR}.

\begin{figure*}[t]
        \centering
        \includegraphics[width=.9\linewidth]{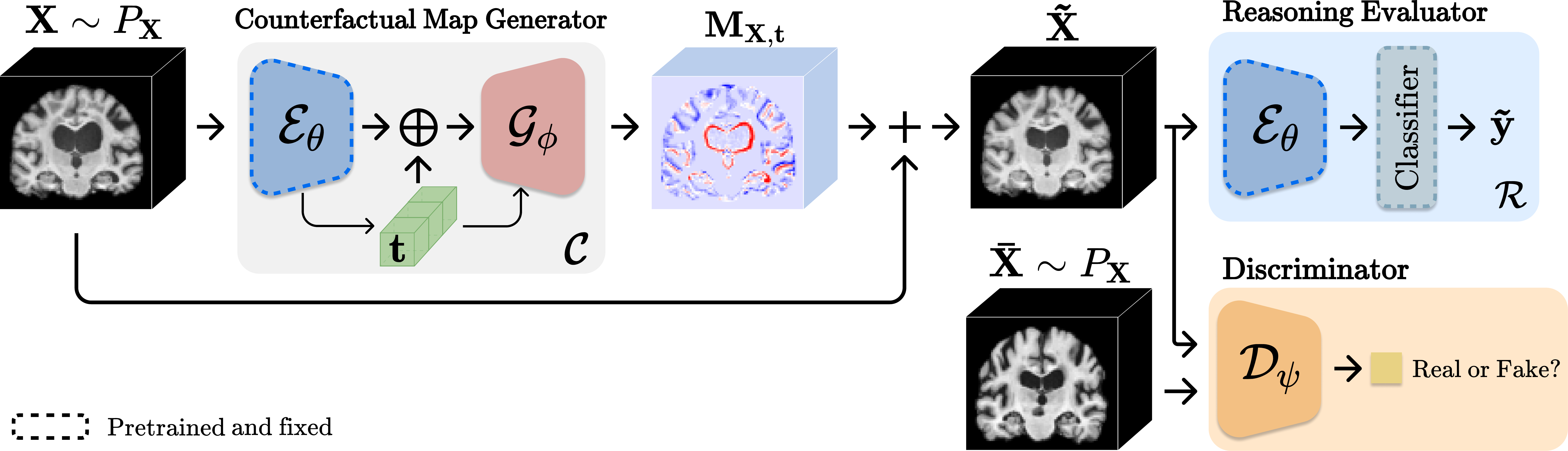}
        \caption{Schematic overview of the counterfactual map generation to induce the cause of dementia diagnosed from the backbone network. It has major components: counterfactual map generator and reasoning evaluator. The counterfactual map generator synthesizes a counterfactual map conditioned on arbitrary target label $\tar$ or posterior probability $\tar'$ obtained from the diagnostic model $\mathcal{R}(\x)$, while the reasoning evaluator works towards enforcing target label attributes to the synthesized map. Note that $\oplus$ is the operator for channel-wise concatenation and $+$ is the operator for element-wise addition.}
        \label{fig:BIN}
\end{figure*}

\section{Related Work}\label{sec2}
In this section, we describe various works proposed for explainable AI (XAI) and attention with guidance approaches for the improvement of AD diagnosis using sMRI.

\subsection{Attribution-based Explanations}
     Attribution-based explanation refers to discovering or identifying the regions that have the most influence on deriving the outcome of a model. The methodological approaches for attribution-based explanation can be subdivided into gradient-based methods and reference-based methods. A gradient-based method highlights the activation nodes that contributed the most to a model’s output. For example, class activation map (CAM)~\cite{zhou2016learning}, and Grad-CAM~\cite{selvaraju2017grad} highlight the activation patterns of weights in a specified layer. Similarly, DeepTaylor~\cite{montavon2017explaining}, DeepLift~\cite{shrikumar2017learning}, and layer-wise relevance propagation (LRP)~\cite{bach2015pixel} pinpoint the attributes that contributed to a model's output score by tracing back via gradient-based computations. These methods usually suffer from vanishing gradients especially when using ReLU activation. Integrated Gradients~\cite{sundararajan2017axiomatic} resolves this issue through sensitivity analysis. However, a crucial drawback of gradient-based methods is that they tend to ignore features with a relatively low discriminative power or to highlight only those with overwhelming feature importance. Furthermore, in general, the highlighted or pinpointed attributes need a secondary analysis or interpretation for human-level understanding. For example, when voxels in the subcortical regions of an input MRI are localized as the informative attributes for MCI/AD identification, it is necessary to further analyze whether those regions involve atrophic changes or morphological variations, which can only be done by experts. 
     
     Reference-based explanation methods~\cite{zeiler2014visualizing,fong2017interpretable,chang2018explaining,dhurandhar2018explanations} focus on changes in model output with regards to perturbation in input samples. Various perturbation methods that employ strategies such as masking~\cite{dabkowski2017real}, heuristics~\cite{chang2018explaining} (\eg, blurring and random noise), using the region of the distractor image as reference for perturbation~\cite{goyal2019counterfactual}, and synthesized perturbation~\cite{dhurandhar2018explanations,zeiler2014visualizing,fong2017interpretable}, have been introduced in the literature. One general drawback of these aforementioned attribution-based explanations is that they tend to produce a similar or blurred salient map (\ie, coarse-grained maps) owing to the unfavorable phenomenon of attribution vanishing.
     
\subsection{Counterfactual Visual Explanations}
     Recently, more researchers have focused on counterfactual reasoning as a form of visual explanation. Counterfactual explanation refers to analyzing a model's output with regard to hypothetical scenarios. For example, in AD diagnosis, a counterfactual explanation could highlight brain regions that may (hypothetically) cause a normal subject to be diagnosed with a disease when transforming an input image accordingly. VAGAN~\cite{baumgartner2018visual} uses a variant of GAN to synthesize a counterfactual map that transforms an input sample to be classified as another label. However, VAGAN has considerable limitations in its framework. First, for map generation, the true label of an input sample must be known, which is not practically possible in real-world scenarios. Second, VAGAN performs a single-way synthesis only. That is, it generates a counterfactual map that transforms an input originally classified as `A' to be classified as `B', but not the reverse. Circumventing the major limitations of VAGAN described above, Bass \etal~proposed ICAM~\cite{bass2020icam} for producing dual-way counterfactual explanations. However, it cannot be used in tasks with multiple classes of interest and is restricted to being a dual-way explanation. 
     
     It is also noteworthy that VAGAN and ICAM focus on generating images to be classified as another specified target class, rather than elucidating the reasoning or explaining a classifier's decision.
     In this work, we propose a novel counterfactual explanation method that can be differentiated from the aforementioned methods as follows:
    (1) Our proposed method is fundamentally designed to generate counterfactual maps to explain a predictive model's output in a post-hoc manner.
    (2) Our proposed method is applicable to a predictive model trained for multi-class classification tasks, \eg, CN \textit{vs.} MCI \textit{vs.} AD and handwritten digit recognition (MNIST), in generating \textit{multi-way} counterfactual maps in a single framework.
    (3) Our proposed LEAR framework is designed to work with most connectionist models, such as ResNet18~\cite{he2016deep}, VoxCNN~\cite{korolev2017residual}, and SonoNet16~\cite{baumgartner2017sononet}, for generating counterfactual maps.
    

\subsection{Attention with Guidance}
Inspired by the recent successes of deep learning techniques using anatomical landmarks in sMRI, several studies have utilized deep neural networks to guide anatomically and neurologically meaningful regions for brain disease diagnosis~\cite{lian2018hierarchical,li2019novel,lian2020attention}. Lian~\etal\cite{lian2018hierarchical} proposed a hierarchical fully convolutional network (H-FCN) using anatomical landmarks, which were used as prior knowledge to rule out non-disease-related regions via an attention mechanism, so as to learn discriminative representations more efficiently. The attention-guided HybNet~\cite{lian2020attention} was also proposed to extract discriminative patches and regions from a whole-brain MRI by exploiting CAM extracted from pre-trained models, upon which multi-scale features were jointly trained and fused to construct a hierarchical classification model for AD diagnosis. In the same line of strategies, Li~\etal~\cite{li2019novel} proposed an iterative guidance method using the CAM for joint pathological region localization and identification for enhancing the diagnostic performance.

While the AD-induced anatomical changes in a brain are subtle, especially in the preclinical or prodromal stages, and are heterogeneous across patients, the aforementioned CAM-based methods can take advantage of only coarse-grained guidance because of the blurry nature of CAM. By contrast, the counterfactual maps obtained from our visual explanation method can provide fine-grained guidance as they represent the minimal source of information to change the clinical label of an input MRI into other ones. 
By regarding the counterfactual maps as \emph{privileged information}, we devise a novel explanation-guided attention (XGA) module
that helps reinforce the generalizability of the predictive network, thus improving its diagnostic performance.

In an effort to categorize the related works of our article, we have categorized some state-of-the-art works into visual explanation, use of attention mechanism, explanation-guided methods, and the ability to reinforce visual explanation, as presented in Table~\ref{tab:rel_works}. Our work is, to the best of our knowledge, the first that exploits an explanation output to improve the generalization of a diagnostic model reciprocally.


\section{Method}\label{sec3}
In this section, we describe our LEAR framework for visual explanation and reinforcing a diagnostic model.
As schematized in Fig. \ref{fig:overview}, there are two principal units in the framework. The first is an \emph{explanation unit} (EU) that learns a counterfactual map generator $\mathcal{C}$, aimed to visually explain the output of a pre-trained diagnostic model for AD/MCI/CN diagnosis. The other one is a \emph{reinforcement unit} (RU) that, guided by the counterfactual maps generated in EU, updates the learnable parameters of the diagnostic model to improve its generalizability and performance. In addition to these two principal units, our framework also involves a step of pre-training a diagnostic model in a conventional manner, \ie, supervised learning using training samples. 

Throughout this article, we denote network models including a diagnostic model, a counterfactual map generation model, and their subnetworks using calligraphic font, while vectors and matrices are denoted by boldface lower and uppercase letters, respectively. The sets are denoted using a typeface style.

Without loss of generality, we assume that a diagnostic model $\mathcal{R}$ is a CNN-based architecture (\eg, ResNet18~\cite{he2016deep}, VoxCNN~\cite{korolev2017residual}, and SonoNet16~\cite{baumgartner2017sononet}) and is trained using a whole-brain 3D MRI as input.

\subsection{Counterfactual Visual Explanation Model}
    \label{counterfactual explanations}
Given a pre-trained diagnostic model $\mathcal{R}$, we describe our novel visual explanation model $\mathcal{C}$ for the output of the diagnostic model. Formally, the goal of our visual explanation model $\mathcal{C}$ is to infer a counterfactual reasoning map over the output label from a diagnostic model. To this end, we develop a counterfactual map generation method in a GAN framework. 

The overall structure for learning our visual explanation model is illustrated in Fig. \ref{fig:BIN}. It has three major modules of a counterfactual map generator (CMG), a reasoning evaluator (RE), and a discriminator (DC). The role of the three modules can be summarized as follows:
\begin{itemize}
	\item CMG: Given an input MRI sample $\mathbf{X}$ and \ks{a target label $\mathbf{t}$, where 
	$\tar=[0,1]^{|\mathcal{Y}|}$
	and $|\mathcal{Y}|$ is the size of the class space $\mathcal{Y}$,} CMG generates a map $\mathbf{M}_{\mathbf{X},\mathbf{t}}$ which, when added to the input $\mathbf{X}$, \ie, $\tilde{\mathbf{X}}=\mathbf{X}+\mathbf{M}_{\mathbf{X},\mathbf{t}}$, causes the transformed image $\tilde{\mathbf{X}}$ to be categorized into the target label $\mathbf{t}$ with high confidence.
	\item RE: This basically exploits the diagnostic model $\mathcal{R}$ itself. It directly evaluates the effect of the generated counterfactual map $\mathbf{M}_{\mathbf{X},\mathbf{t}}$ in producing the targeted label $\mathbf{t}$, possibly diagnosed differently from the output label of the original input $\mathbf{X}$.
	\item DC: This helps the CMG to generate an anatomically and morphologically meaningful map, making the transformed image $\tilde{\mathbf{X}}$ realistic.
\end{itemize}
As RE and DC are, respectively, the network of a diagnostic model and a typical component in GAN, we describe only the CMG in detail. 
    
    \begin{figure}[t]
        \centering
        \includegraphics[width=.95\linewidth]{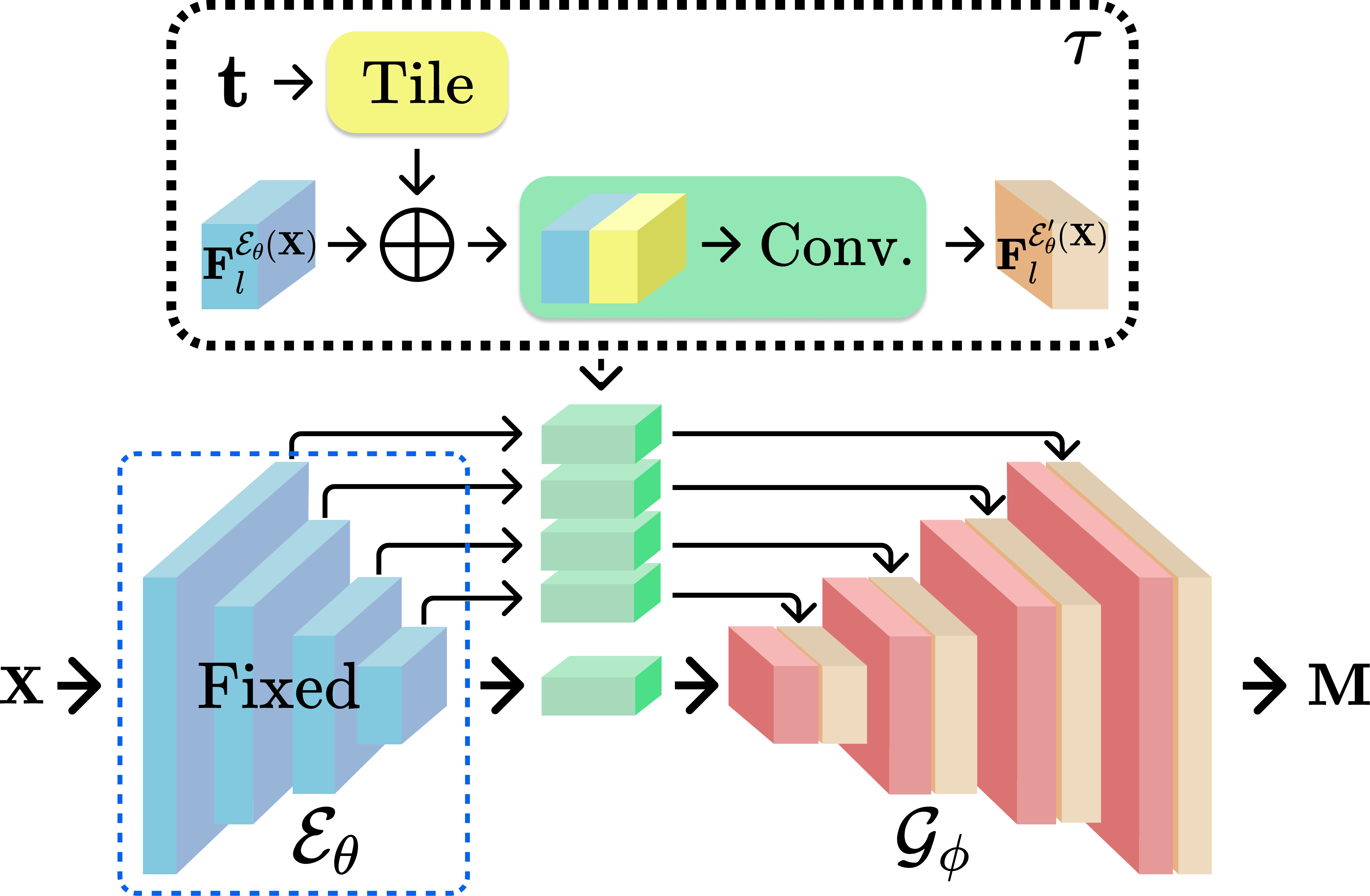}
        \caption{Detailed view of the counterfactual map generator (CMG). A target label $\tar$ is tiled and channel-wise concatenated to the skip connection. This enables the CMG to condition the counterfactual maps to be conditioned on an arbitrary target condition.}
        \label{fig:generator skip details}
    \end{figure}
    
    \subsubsection{Counterfactual Map Generator (CMG)}
    The CMG is a variant of Conditional GAN~\cite{mirza2014conditional} that can synthesize a counterfactual map $\M_{\x,\tar}$ conditioned on a target label $\tar$. It consists of an encoder $\mathcal{E}_\theta$ and a generator $\mathcal{G}_\phi$, where the subscripts $\theta$ and $\phi$ denote the tunable parameters of the respective networks. The network design of the encoder $\mathcal{E}_\theta$ and the generator $\mathcal{G}_\phi$ is a variant of U-Net~\cite{ronneberger2015u} with a tiled target label concatenated to the skip connections, as presented in Fig.~\ref{fig:generator skip details}. Here, we should emphasize that the encoder $\mathcal{E}_{\theta}$ is taken from the set of layers and the corresponding parameters to extract features in a pre-trained diagnostic model $\mathcal{R}$ with weights $\theta$ fixed. Therefore, the encoder $\mathcal{E}_{\theta}$ is already capable of extracting disease-related features from an input $\mathbf{X}$, thus making our CMG trainable relatively easily and robustly by tuning the parameters of layers other than in the encoder $\mathcal{E}_{\theta}$ only.
    
    \ks{Let $\{\mathbf{F}_l^{\mathcal{E}_{\theta}(\x)} \}^L_{l=1}$ denote the output feature maps of the $L$ convolution layers in the encoder $\mathcal{E}_{\theta}(\x)$.} A given target label $\tar$ is concatenated with the feature maps after tiling so that their shapes match the respective feature maps concatenated, \ks{\ie, tile $\mathbf{t}$ with the size of $w_l\times h_l\times d_l\times c$ where $w_{l}$, $h_{l}$, and $d_{l}$ denote, respectively, the width, height, and depth of a feature map from the $l$-th convolution block.} In order to extract better representations of the target label related information, we apply a convolution operation ($\operatorname{Conv3D}$) with a learnable $3\times 3\times 3$ kernel, a stride of 1 in each dimension, and zero padding, followed by a nonlinear LReLU activation function as follows (see Fig.~\ref{fig:generator skip details}): 
    \begin{equation}
        \ks{\tau(\mathbf{F}_{l}^{\mathcal{E}_{\theta}(\x)},\tar) = \operatorname{LReLU}\left(\operatorname{Conv3D}\left(\mathbf{F}_{l}^{\mathcal{E}_{\theta}(\x)}\oplus\operatorname{Tile}(\mathbf{t})\right)\right)}
    \end{equation}
    where $\oplus$ denotes an operator of channel-wise concatenation.
Then, the target label information included feature maps $\mathbf{F}_{l}^{\mathcal{E}_{\theta}'}=\tau(\mathbf{F}_{l}^{\mathcal{E}_{\theta}(\x)},\tar)$ are transmitted to the generator $\mathcal{G}_{\phi}$ via skip connections.
The generator $\mathcal{G}_{\phi}$ is then able to responsibly synthesize a map $\M_{\x,\tar}$ from the target label informed feature maps as follows:
        \begin{equation}
            \ks{\M_{\x,\tar}=\mathcal{G}_\phi\left(\mathcal{T}(\x,\tar)\right)}
            \label{cm_map_original}
        \end{equation}
\ks{where $\mathcal{T}(\x,\tar)=\{\tau(\mathbf{F}_{1}^{\mathcal{E}_{\theta}(\x)},\tar),...,\tau(\mathbf{F}_{L}^{\mathcal{E}_{\theta}(\x)},\tar)\}$.}
Finally, we produce a transformed MRI $\tx$ by combining the synthesized map $\M_{\x,\tar}$ with an input MRI $\mathbf{X}$ via addition, \ie, $\tx=\x+\M_{\x,\tar}$, which is supposed to be classified as the target label $\mathbf{t}$ by the following RE module, \ie, the diagnostic model $\mathcal{R}$.

Note that in a setting where the target label $\tar$ (\eg, CN) is different from the ground-truth label $\y$ (\eg, AD), 
 we may hypothesize that the synthesized map $\M_{\x, \tar}$ visually explains why the input $\x$ was classified to $\tar_{\x}$ (\eg, AD), instead of $\tar$ (\eg, CN) because $\M_{\x, \tar}$ highlights the hypothetical regions that contributed to transforming an AD-like MRI $\x$ to a CN-like MRI $\tx$.

\subsubsection{Counterfactual Visual Explanation Model Training}
In this subsection, we define a set of loss functions to train our counterfactual visual explanation model.

\noindent{\bfseries Cycle Consistency}: 
{\HI In order to encourage the synthesized map $\M_{\x,\tar}$, which is conditioned on an input $\x$ and a target label $\tar$, to be anatomically and morphologically meaningful, we exploit a cycle consistency loss \cite{zhu2017unpaired} with $\ell_{1}$-norm as follows:}
\ks{\begin{gather}
        \begin{aligned}
            \mathcal{L}_{\text{cyc}} = \mathbb{E}_{\x\sim P_{\x},\tar\sim U(0,|\mathbf{y}|)}\left[\left\|\x' - \x\right\|_1\right]
        \end{aligned}
        \label{eq_cyc}
\end{gather}}
    where $P_{\x}$ denotes a distribution of MRI samples, \ks{$|\mathcal{Y}|$ is the number of classes,
    $U(\cdot)$ is the one-hot encoded form of a discrete uniform distribution
    , $\tx=\x+\M_{\x,\tar}$ and ${\x}' = \tilde{\x}+\M_{\tilde{\x},\mathcal{R}(\x)}$.}
As we propose a way of generating multi-way counterfactual maps, this loss is imperative to synthesize different counterfactual maps for different conditions without suffering from a mode collapse problem~\cite{goodfellow2016nips}. 

Note that, in following equations, we omit arbitrary target labels \ks{$\tar\sim U(0,|\mathcal{Y}|)$} from the expectation terms for simplicity.

\noindent{\bfseries Adversarial Learning}: Inspired by Least Square GAN~\cite{mao2017least}, we adopt the least squares loss function that penalizes samples distant from the discriminator's decision boundary.
Using the cycle consistency loss in Eq.~(\ref{eq_cyc}), the least squares loss needs to be applied to real MRI samples $\bar{\x}$, and transformed (\ie, fake) samples $\tx$ and $\x'$.

\begin{align}
	\begin{split}
	 \mathcal{L}_{\text{adv}}^{\mathcal{D}_{\psi}} &= \mathbb{E}_{\bar{\x}\sim P_{\x}}\left[(D_{\psi}(\bar{\x})-1)^2\right]  \\
	 &+ \frac{1}{2}\left(\mathbb{E}_{\x\sim P_{\x}}\left[D_{\psi}(\tx)^2+D_{\psi}(\x')^2\right]\right)
	 \end{split}
	 \label{adv d loss}
	\\
	\begin{split}
	\mathcal{L}_{\text{adv}}^{\mathcal{G}_{\phi}} &= \frac{1}{2}\left(\mathbb{E}_{\x\sim P_{\x}}\left[(D_{\psi}(\tx) - 1)^2+(D_{\psi}(\x') - 1)^2\right]\right)
        \end{split}\label{adv g loss}
\end{align}
    This objective function is very suitable for our CMG training because the generated counterfactual maps should neither destroy the input appearance nor ignore the target attribution.

\noindent{\bfseries Total Variation}: 
For a more natural synthesis of the counterfactual map generated from CMG and its harmonization with an input sample, we exploit the total variation loss~\cite{gottlieb1998total} as a regularizer.
\begin{align}
    \begin{split}
        	\mathcal{L}_{\text{tv}} = \sum_{i,j,k}\left|\tilde{\x}_{i+1,j,k}-\tilde{\x}_{i,j,k}\right| &+ \left|\tilde{\x}_{i,j+1,k}-\tilde{\x}_{i,j,k}\right|\\ &+ \left|\tilde{\x}_{i,j,k+1}-\tilde{\x}_{i,j,k}\right|
    \end{split}\label{eq_tv}
\end{align}
where $\tilde{\x}=\x+\M_{\x, \tar}$, and  $i$, $j$, and $k$ are indices of each axis in the 3D coordinate of a volumetric image, respectively.

\noindent{\bfseries Sparsity in a Counterfactual Map}:
{\HI From the interpretability and identity preservation standpoints, it is crucial to regularize the dense counterfactual map to highlight only the essential regions necessary for counterfactual reasoning. To this end, we also impose an elastic regularization on the synthesized counterfactual map as follows:}
    \begin{equation}
            \mathcal{L}_{\text{map}} = {\mathbb{E}_{\x\sim P_{\x}}}\left[\lambda_1\left\|\M_{\x, \tar}\right\|_1 + \lambda_2\left\|\M_{\x, \tar}\right\|_2\right]\label{cm loss}
    \end{equation}
    where $\lambda_1$ and $\lambda_2$ {\HI are the weighting hyperparameters}.

\noindent{\bfseries\HI Correctness of Counterfactual Reasoning}:
    To ensure that the transformed image $\tx=\x+\M_{\x,\tar}$ is correctly conditioned on the target label $\tar$, we include a classification loss function as follows: 
    \begin{equation}
            \mathcal{L}_{\text{cls}} = \mathbb{E}_{\x\sim P_{\x}}\left[\operatorname{CE}\left(\tar,\tilde{\y})\right)\right]\label{cls_loss}
    \end{equation}
    where $\operatorname{CE}$ denotes a cross-entropy function, and $\tilde{\y}=\mathcal{R}(\tx)$ is a softmax activated probability. 
    
    Conceptually, the role of the diagnostic model $\mathcal{R}$ is similar to that of a discriminator $\mathcal{D}_{\psi}$, but their objective is very different. While a discriminator $\mathcal{D}_{\psi}$ learns to distinguish between real and fake samples, the diagnostic model $\mathcal{R}$ is already trained to classify the input samples correctly. Thus, the diagnostic model $\mathcal{R}$ provides a deterministic guidance for the generator to produce a target-directed counterfactual map, while the discriminator $\mathcal{D}_{\psi}$ plays a min-max game with a generator $\mathcal{G}_{\phi}$ in an effort to produce more realistic samples.

\subsubsection {Total Loss Function}
    We define the total loss function for counterfactual map generation as follows:
	\begin{align}
		 \mathcal{L}_{\text{CMG}}=\cmt\lambda_{3}\mathcal{L}^{\mathcal{G}_\phi}_{\text{adv}}+\lambda_{4}\mathcal{L}^{\mathcal{D}_\psi}_{\text{adv}}+\lambda_{5}\mathcal{L}_{\text{cyc}}+\lambda_{6}\mathcal{L}_{\text{cls}}+\lambda_{7}\mathcal{L}_{\text{tv}}+\mathcal{L}_{\text{map}},\label{total_cmg_loss}
	\end{align}
    {\cmt where $\lambda_{*}$ values are the hyperparameters of the model ($\lambda_{1,2}$ in Eq.~(\ref{cm loss}))}.
    
It should be noted that during training, we share and fix the weights of the encoder $\mathcal{E}_\theta$ of the CMG with the RE module's feature extractor $\mathcal{E}_\theta$ to ensure that the attribution is consistent throughout the generative process.

\begin{figure*}[t]
    \centering
    \includegraphics[width=1.\linewidth]{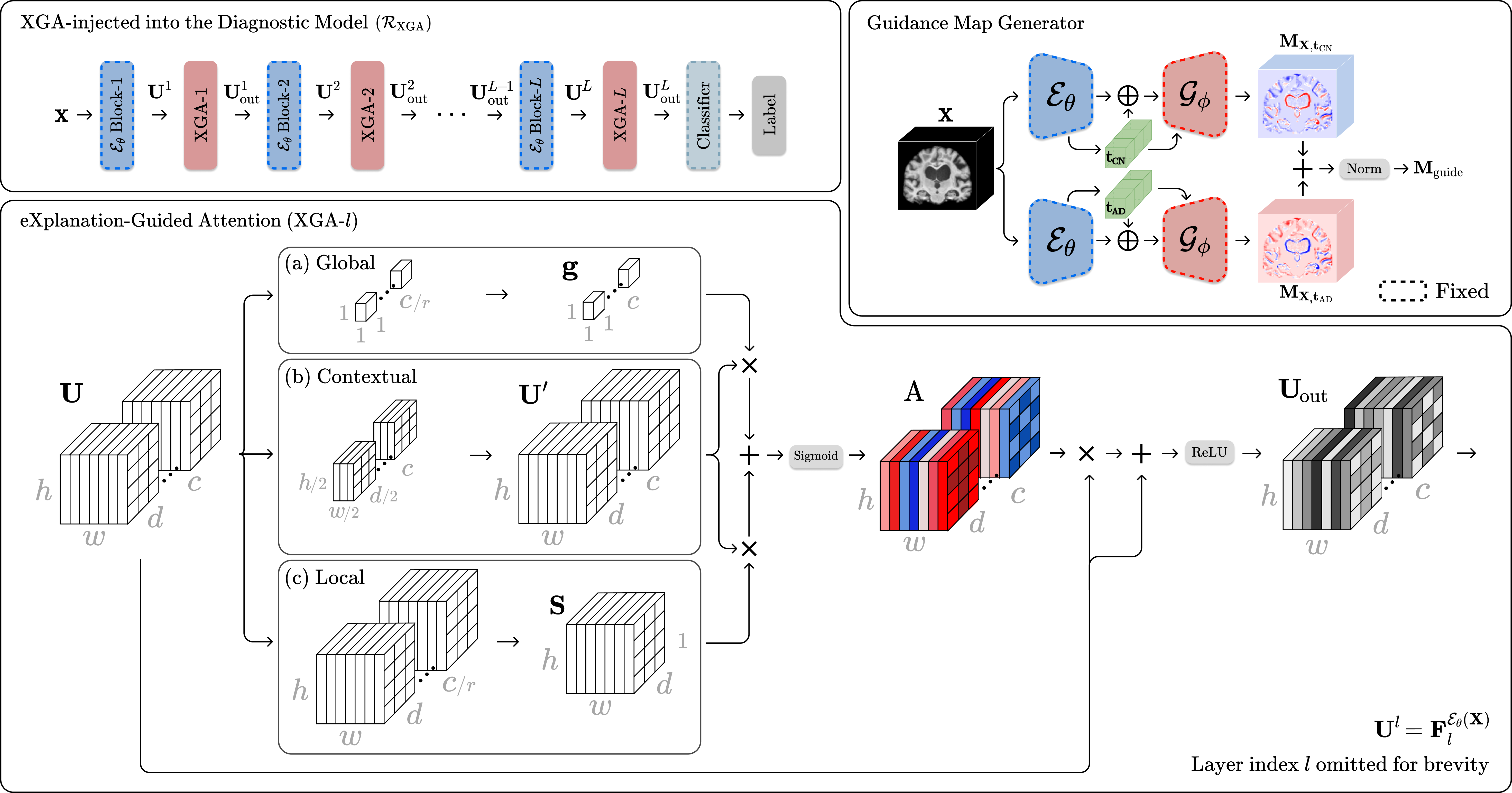}
    \caption{Schematic overview of the explanation-guided attention (XGA) module with a guidance map. A guidance map is a supervision for the XGA module that assists in focusing on regions of pathological and morphological changes caused by dementia on the whole-brain. XGA module learns and integrates locally subtle changes and globally discriminative structural changes that can optionally be supervised by the guidance map. Note that $\times$ is the operator for element-wise product and $+$ is the operator for element-wise addition.}
    \label{fig:attention}
\end{figure*}
    
\subsection{\HI Reinforcement Representation Learning}
\label{guidance maps}

In this article, we hypothesize that the set of counterfactual maps synthesized by our CMG along with a diagnostic model can be a vital source of knowledge of anatomical or morphological changes relevant to AD, inferred in a data-driven manner. Such data-driven knowledge is comparable to the conventional neuroscientific knowledge mostly acquired from a group statistical analysis in a univariate manner \cite{frisoni2010clinical}. Note that the diagnostic model is trained with the aim of classifying samples having different clinical labels, \eg, CN, MCI, AD, by discovering generalizable and discriminative patterns inherent in samples. Our proposed CMG is designed and trained to detect such generalizable and discriminative patterns in an input sample to explain the diagnostic model's output via counterfactual reasoning.

\subsubsection{Guidance Map Generation}\label{sec:GMG}

Based on these considerations, we propose to exploit the counterfactual maps as guidance to reinforce the diagnostic model's representations. Specifically, we generate the counterfactual maps of an input sample with the target labels of $\tar_{\text{CN}}$ and $\tar_{\text{AD}}$, \ie, the most normal and the most AD-like brains with regard to the input sample. For example, in a 3-class classification task of CN \textit{vs.} MCI \textit{vs.} AD, $\tar_{\text{CN}}=[1,0,0]$ is the one-hot vector of CN and $\tar_{\text{AD}}=[0,0,1]$ is the one-hot vector of AD.
Assuming that these two counterfactual maps jointly represent AD-sensitive brain regions globally and locally, we build a guidance map $\M_{\text{guide}}$ by combining them as follows:
\begin{align}
    \M_{\text{guide}} = \operatorname{MinMax}\left(\left|\M_{\x,\tar_{\text{CN}}}\right| + \left|\M_{\x,\tar_{\text{AD}}}\right|\right)
    \label{guide_map}
\end{align}
where $\left|\cdot\right|$ is an absolute operation and $\operatorname{MinMax}(\cdot)$ denotes a min-max normalization in a voxel-wise manner.
\ks{Thus, the absolute term in the guidance map $\M_\text{guide}$ allows the use of attentive values in both the extreme cases of most normal brain and most AD-like brain (because negative values of $\M_{\x,\tar_\text{CN}}$ highlight the most normal regions of the brain, while positive values of $\M_{\x,\tar_\text{AD}}$ highlight the most AD-like regions of the brain).}
This guidance map is then used to reinforce the representational power of the layers' outputs in the diagnostic model by modulating them via the attention mechanism described below.

\subsubsection{Explanation-Guided Attention}
In order to exploit the explanation-induced knowledge of the anatomical and morphological changes for AD diagnosis, we devise an explanation-guided attention (XGA) module by regarding the counterfactual maps as \emph{model-driven privileged information} during training. Specifically, we inject a self-attention module that adaptively modulates the layer outputs in the diagnostic model (Fig. \ref{fig:attention}). 

Let $\mathbf{U}^{l}$ be an output feature map of the $l$-th layer in the diagnostic model $\mathcal{R}$, \ie, $\mathbf{U}^{l}=\mathbf{F}_l^{\mathcal{E}_\theta}$, and $\mathbf{A}^{l}$ its resulting attention map, whose computation is detailed below. 
\ks{Note, it is expected that the attention map $\mathbf{A}^{l}$ produces the higher attentive values, where the higher explanation values are in the guidance map $\M_{\text{guide}}$, obtained by Eq.~(\ref{guide_map}), for an input sample.}
Thereby, the AD-sensitive regions, guided by the counterfactual maps, are excited with the discriminative representations while other regions are inhibited, thus reinforcing the feature representations in the diagnostic model. 

We base our computation to estimate the attention map on the global-and-local (GALA) module~\cite{linsley2018learning}, \ks{which consists of global and local operators, as presented in Fig. \ref{fig:attention} (a), (c), respectively.
Specifically, our XGA module adapts the local and global attention operators of GALA and improves them using a contextual attention operator. Using this simple modification to the GALA attention mechanism, our XGA module achieved about 11\% accuracy improvement in 3-class diagnosis experiments {\supp{(results in Supplementary S5).}}}

Basically, the following operations can be applied to different layers equally. Hereafter, we omit the superscript $l$ of a layer index to reduce clutter.
Our XGA modulates an input feature map $\mathbf{U}$ with an attention map $\mathbf{A}$ of the same dimension as $\mathbf{U}$. That is, $\mathbf{U},\mathbf{A}\in\mathbb{R}^{w\times h\times d\times c}$, where $w$, $h$, $d$, and $c$ are the spatial width, height, depth, and number of feature channels, respectively.

\noindent{\bfseries Global Attention:} First, we account for the global attention in the XGA module by exploiting the squeeze-and-excitation technique~\cite{linsley2018learning,hu2018squeeze}. To obtain the global feature attention vector $\mathbf{g}\in\mathbb{R}^{1\times1\times1\times C}$, we first obtain a channel descriptor $\mathbf{d}=\left[d_{c}\right]_{c=1}^{C}$ by calculating the summary statistics of the $c$-th channel via global average pooling, \ie, $d_{c}=\frac{1}{WHD}\sum^{W}_{w=1}\sum^{H}_{h=1}\sum^{D}_{d=1}{U}_{whdc}$.
As the channel descriptor $\mathbf{d}$ includes information obtained from the full receptive field, it can be thought of as carrying the importance of the respective channel with respect to the global information. This is followed by a two-layer neural network that non-linearly transforms the channel descriptor to explicitly model the inter-dependencies among the channels as follows:
\begin{align}
    \mathbf{g} = \mathbf{W}_{\text{expand}}(\operatorname{ReLU}(\mathbf{W}_{\text{c-shrink}}(\mathbf{d}))).
\end{align}
where $\mathbf{W}_{\text{c-shrink}}\in\mathbb{R}^{\frac{C}{r}\times C}$ and $\mathbf{W}_{\text{expand}}\in\mathbb{R}^{C \times \frac{C}{r}}$ are the shrinking and expansion operations, respectively, and $r$ is a ratio hyperparameter.

\noindent{\bfseries Local Attention:} Second, we consider a local saliency component to compute the local feature attention $\mathbf{S}$. Unlike the global attention, the local feature attention map $\mathbf{S}$ focuses on ``where'' a crucial part locates, complementing the global attention. While retaining the spatial dimensions, we conduct two consecutive convolution operations along the channel dimension with a non-linear transformation in-between to enhance the complexity as follows:
\begin{align}
    \mathbf{S} = \mathbf{W}_{\text{collapse}}*(\operatorname{ReLU}(\mathbf{W}_{\text{d-shrink}}*\mathbf{U}))
\end{align}
where $*$ denotes convolution, $\mathbf{W}_{\text{d-shrink}}\in\mathbb{R}^{1\times 1\times 1\times C\times\frac{c}{r}}$ and $\mathbf{W}_{\text{collapse}}\in\mathbb{R}^{1\times 1\times 1\times \frac{c}{r}\times 1}$ are learnable parameters. This local attention is used to generate a spatial attention map by allowing for the inter-spatial relationship of features.

\noindent{\bfseries Contextual Attention:} \ks{Along with the global and local attention operators of GALA described above}, our XGA also involves a contextual attention operator. It is designed to utilize the contextual information from a larger receptive field. To this end, we first conduct a dilated convolution~\cite{chen2017deeplab}, which has the effect of taking into account features of enlarged field of view and reducing the map size, followed by a non-linear transformation. We then up-scale its output back to the input size of $\mathbf{U}$ as follows:
\begin{align}
    \mathbf{U}' = \operatorname{Up}(\operatorname{ReLU}(\mathbf{W}_{\text{reduction}} *_{d} \mathbf{U}))
\end{align}
where $*_{d}$ denotes dilated-convolution and $\operatorname{Up}(\cdot)$ is an operator for trilinear up-scaling back to the original spatial dimensions of $\mathbf{U}$.

Finally, the global, local, and contextual module outputs are integrated to produce the attention mask $\mathbf{A}\in\mathbb{R}^{w\times h\times d\times c}$ after tiling $\mathbf{g}$ and $\mathbf{S}$ to form $\mathbf{G}^*$, $\mathbf{S}^* \in \mathbb{R}^{w\times h\times d\times c}$ owing to their differences in size, as follows:
\begin{align}
    \mathbf{A} = \sigma\left(\mathbf{G}^*\times\mathbf{U}' + \mathbf{S}^*\times\mathbf{U}'\right)
\end{align}
where $\sigma$ denotes a sigmoid activation function, $\times$ denotes element-wise product, and $+$ denotes element-wise addition.
The attention mask $\mathbf{A}$ plays the role of excitation and inhibition of the input feature map $\mathbf{U}$ with a skip connection as follows:
\begin{align}
    \mathbf{U}_{\text{out}} = \operatorname{ReLU}(\mathbf{U}+(\mathbf{U}\times\mathbf{A})).
\end{align}

\subsubsection{XGA Learning}
Inspired by Linsley~\etal~\cite{linsley2018learning}, we define the loss using the cross-entropy function regularized by the attention-guidance penalty \ks{$\Omega_{\text{XGA}}$} to train the parameters ${\mathcal{R}_\omega=\{\mathbf{W}_{\text{c-shrink}}^{l}, \mathbf{W}_{\text{expand}}^{l}, \mathbf{W}_{\text{collapse}}^{l}, \mathbf{W}_{\text{d-shrink}}^{l}, \mathbf{W}_{\text{reduction}}^{l}\}_{l=1}^{L}}$ for XGA modules injected next to the every convolution layer of the diagnostic model as follows:
\begin{align}
	\mathcal{L}_{\text{XGA}} = \mathbb{E}_{\x \sim P_{\x}}[\text{CE}(\mathbf{y}, \mathcal{R}_{\text{XGA}}(\x))] + \ks{\lambda_{8}\Omega_{\text{XGA}}}
\end{align}
\begin{align}
	\ks{\Omega_{\text{XGA}}} = \sum_{l  \in L}\left\|\frac{\bar{\mathbf{M}}^{l}_{\text{guide}}}{\|\bar{\mathbf{M}}^{l}_{\text{guide}}\|_{2}}-\frac{\bar{\mathbf{A}}^{l}(\mathbf{X})}{\left\|\bar{\mathbf{A}}^{l}(\mathbf{X})\right\|_{2}}\right\|_{2}
\end{align}
where \ks{$\Omega_{\text{XGA}}$} is a scalar value for attention-guidance penalty, $\mathcal{R}_{\text{XGA}}$ denotes the diagnostic model $\mathcal{R}$ with XGA modules injected, $\bar{\mathbf{A}}^{l}(\mathbf{X})\in \mathbb{R}^{w\times h\times d\times 1}$ is the compression of $\mathbf{A}^{l}(\mathbf{X})\in \mathbb{R}^{w\times h\times d\times c}$ with channel-wise $\ell_{2}$-norm values, $\bar{\mathbf{M}}^{l}_{\text{guide}}$ is a trilinear-interpolated form of $\mathbf{M}^{l}_{\text{guide}}$ to be the same size of $\bar{\mathbf{A}}^{l}(\mathbf{X})$, and $\lambda_{8}$ is a weighting hyperparameter. 
While training the parameters of the XGA modules, we fix the other model parameters of the diagnostic model. 
With regards to the attention-guidance penalty \ks{$\Omega_{\text{XGA}}$}, as described above, we expect that the attention map $\mathbf{A}^{l}(\x)$ of an input sample $\x$ outputs higher values for excitation, where the higher explanation values are in the guidance map $\M_{\text{guide}}$, and lower values for inhibition otherwise.

\subsection{Iterative Explanation-Reinforcement Learning}
Finally, we introduce an iterative explanation-reinforcement learning scheme that enhances the quality of visual explanation as well as the performance of the diagnostic model as follows:

\noindent{\bfseries Phase 1: CMG training} 
\begin{align}
    \min_{\mathcal{G}_\phi}{\mathcal{L}_\text{CMG}}, 
    \min_{\mathcal{D}_\psi}{\mathcal{L}_\text{CMG}},
\end{align}
\noindent{\bfseries Phase 2: XGA training}
\begin{align}
    \min_{\mathcal{R}_\omega}{\mathcal{L}_\text{XGA}}.
\end{align}
In this iterative training scheme, Phase 1 and Phase 2 are repeated sequentially. During the first iteration of the optimization, we use the original definition of counterfactual reasoning map $\M_{\x,\tar}$, \ie, Equation~(\ref{cm_map_original}), as XGA has not reinforced the diagnostic model yet. For second and later iterations, we redefine the counterfactual map as follows:
\begin{equation}
    \ks{\M_{\x,\tar}:=\mathcal{G}_\phi\left(\mathcal{T}'(\x,\tar)\right)}
    \label{cm_renew}
\end{equation}
\ks{where $\mathcal{T}'(\x,\tar)=\{\tau(\mathbf{F}_{1}^{\mathcal{E}_{\theta,\omega}(\x)},\tar),...,\tau(\mathbf{F}_{L}^{\mathcal{E}_{\theta,\omega}(\x)},\tar)\}$, and
$\{\mathbf{F}_l^{\mathcal{E}_{\theta,\omega}(\x)} \}^L_{l=1}$ denote the output feature maps of the $L$ convolution layers in the encoder $\mathcal{E}_{\theta,\omega}(\x)$ from the XGA-injected diagnostic model $\mathcal{R}_\text{XGA}$.} Note that parameters of the pre-trained diagnostic model, \ie, $\theta$, is fixed during all phases and all iterations.

\section{Experimental Settings and Results}\label{sec4}
In this section, we (1) analyze and validate the visual explanation results of our counterfactual reasoning map; (2) show the effectiveness of our LEAR framework in reinforcing the diagnostic models; and (3) apply our LEAR framework to baseline and state-of-the-art diagnostic models to demonstrate its portability.

\subsection{Counterfactual Reasoning}

\subsubsection{Toy Example: MNIST Classifier}\label{sec:mnist}
In order to help the readers' understanding of visual explanation method using counterfactual reasoning maps, we present the visual explanation of an MNIST classifier, owing to its intuitiveness.

\begin{figure}[t]
    \centering
    \includegraphics[width=.95\linewidth]{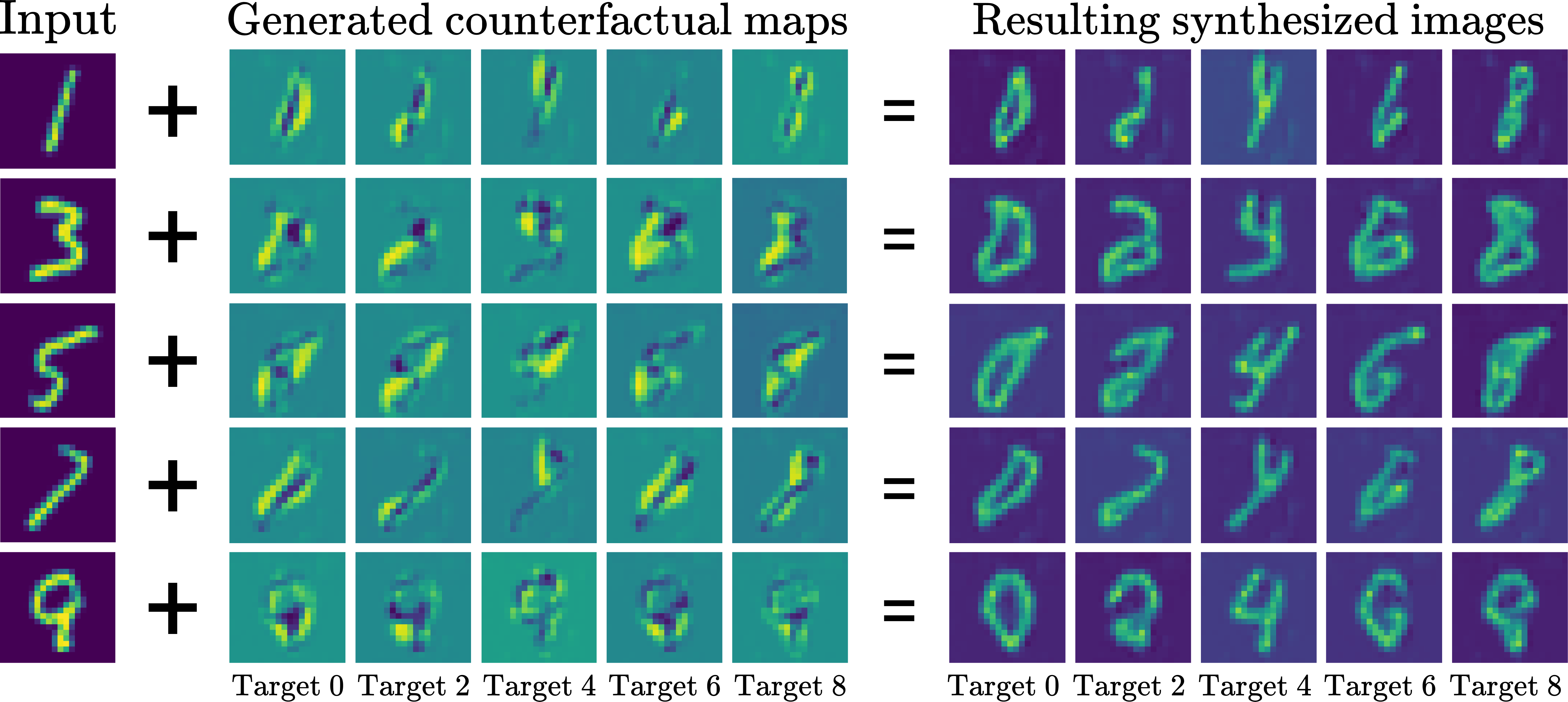}
    \caption{Examples of counterfactual maps for the MNIST dataset. The resulting synthesized image is an addition between an input and its corresponding counterfactual map (blue and yellow denote, respectively, subtraction and addition of the respective pixel values, \ie, deletion and addition of areas to be a target-labeled digit) conditioned on a target label.}
    \label{fig:MNIST}
\end{figure}

\begin{figure*}[t]
    \centering
    \includegraphics[width=.95\linewidth]{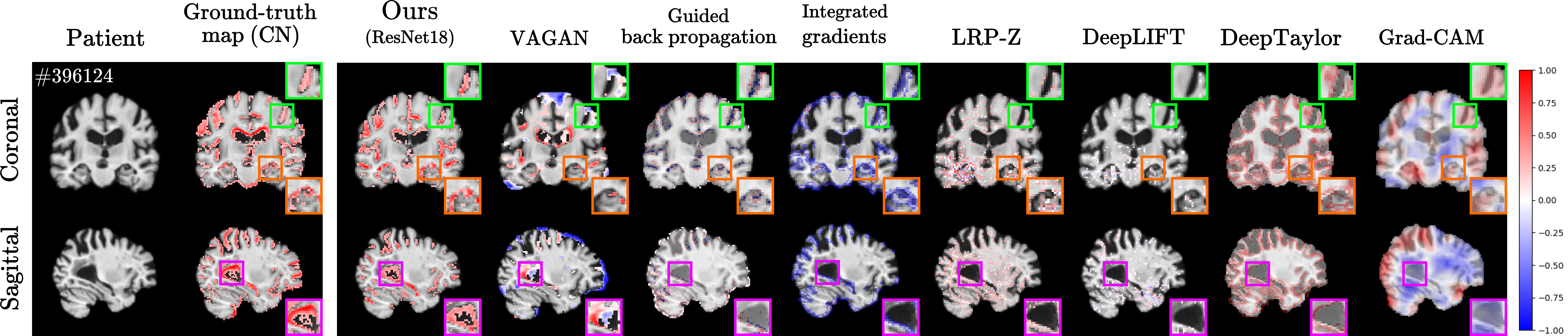}
    \caption{Example of counterfactual maps for the ADNI dataset (Subject ID 024\_S\_0985, Image ID on top left corner). Purple, green, and orange boxes visualize ventricular, cortex, and hippocampus regions, respectively.}\label{fig:Brain_result}
\end{figure*}

\vspace{5pt}\noindent{\bfseries Dataset and Implementation}

\noindent MNIST~\cite{lecun1998mnist} is a gray-scale handwritten digit image dataset that, we believe, is suitable for the proof-of-concept of various visual explanation methods. For the preparation of the dataset, we utilized the data split provided by the dataset publisher~\cite{lecun1998mnist} and applied min-max normalization.
For the classifier model, we re-implemented and pre-trained the model proposed by Kim \etal~\cite{kim2018disentangling} with minor modifications (\eg, kernel and stride size) to accommodate the smaller image size of the MNIST dataset. More details on the implementation are {\supp{in Supplementary S1.1.}}

\vspace{5pt}\noindent{\bfseries Results and Analysis}

\noindent
Fig.~\ref{fig:MNIST} shows examples of the generated counterfactual (CF) maps and the resulting synthesized images towards five targeted classes (\ie, 0, 2, 4, 6, 8) from six different input images.
Note that our CMG successfully produced counterfactual visual explanations, indicating which pixels should be deleted (blue) or added (yellow) to be the different target classes, in \textit{multiple} hypothetical scenarios.

We emphasize the importance of visual explanation in \textit{hypothetical scenarios} as it can provide users with an intuitive understanding on \emph{``what if $\mathbf{X}$ was $\mathbf{X}^*$?''}
In this sense, a CF map should transform an input sample $\x$ to be dependent only on the targeted hypothetical scenario and independent to any other artifacts. Our experiment on the MNIST dataset demonstrates this ability to isolate attribution to only the targeted label because the transformed image maintains the style of the input image while successfully being transformed to a target digit.
For example, for transforming an image of the digit ``3'' to a target digit ``8'', we can observe that the contours of the original ``3'' image are maintained while new contours are added to form a digit ``8''.
Likewise, for transforming an image of the digit ``0'' to a target digit ``4'', we can observe that the upper and bottom arc was removed to form a digit ``4'' with the rest of the arcs maintained.
This ability to isolate targeted conditions allows our CMG module to be reliably applied to a medical task in the next subsection.

\subsubsection{Alzheimer's Disease Classifier}

\vspace{5pt}\noindent{\bfseries Dataset and Implementation}

\noindent The ADNI dataset, collated by the Alzheimer’s Disease Neuroimaging Initiative~\cite{MUELLER2005869}, is used for the following experiments. The ADNI dataset is highly challenging as it is practice-oriented in real-world medical applications and its images feature subtle and diverse morphological changes. It consists of 3D structural magnetic resonance imaging (sMRI) of various subject groups ranging from cognitive normal (CN) to Alzheimer's disease (AD). Specifically, we have utilized 431 CN subjects, 497 stable mild cognitive impairment (sMCI), 251 progressive mild cognitive impairment (pMCI), and 359 AD subjects in ADNI-1 and ADNI-2 studies.
Some subjects had multiple MRIs acquired during the span of their life, but we have only selected their baseline MRIs. Thus, 1,538 images are used in our experiments.
For three-class experiments, sMCI and pMCI subjects were considered as MCI subjects. Note that, in our experiments, prodromal stages from CN to AD are sMCI and pMCI, with the latter generally considered more severe.

We used a five-fold cross-validation setting for all experiments, and used the same indices for all the comparison methods. We made sure there was no data leakage while training the backbone diagnostic models, CMG optimization, XGA optimization, and iterative optimization. 

ResNet18~\cite{he2016deep} baseline and various state-of-the-art diagnostic models were re-implemented for the encoder $\mathcal{E}_\theta$ to demonstrate the generalizability of our LEAR framework. Note that, unless specified otherwise, we utilize the three-class ResNet18 model as the backbone diagnostic model for the experiments in this section. The decoder $\mathcal{G}_\phi$ in the CMG has the same network design as the encoder $\mathcal{E}_\theta$ with pooling layers replaced by up-sampling layers. 
\ks{We have also utilized the structure of encoder $\mathcal{E}_{\theta}$ as the DC module $\mathcal{D}_\psi$ identically in all experiments.}
\ks{More details on the implementation, ADNI dataset, and sMRI preprocessing are provided {\supp{in Supplementary S1.2 and S2.1.}}}

\begin{figure}[t]
    \centering
    \includegraphics[width=1.\linewidth]{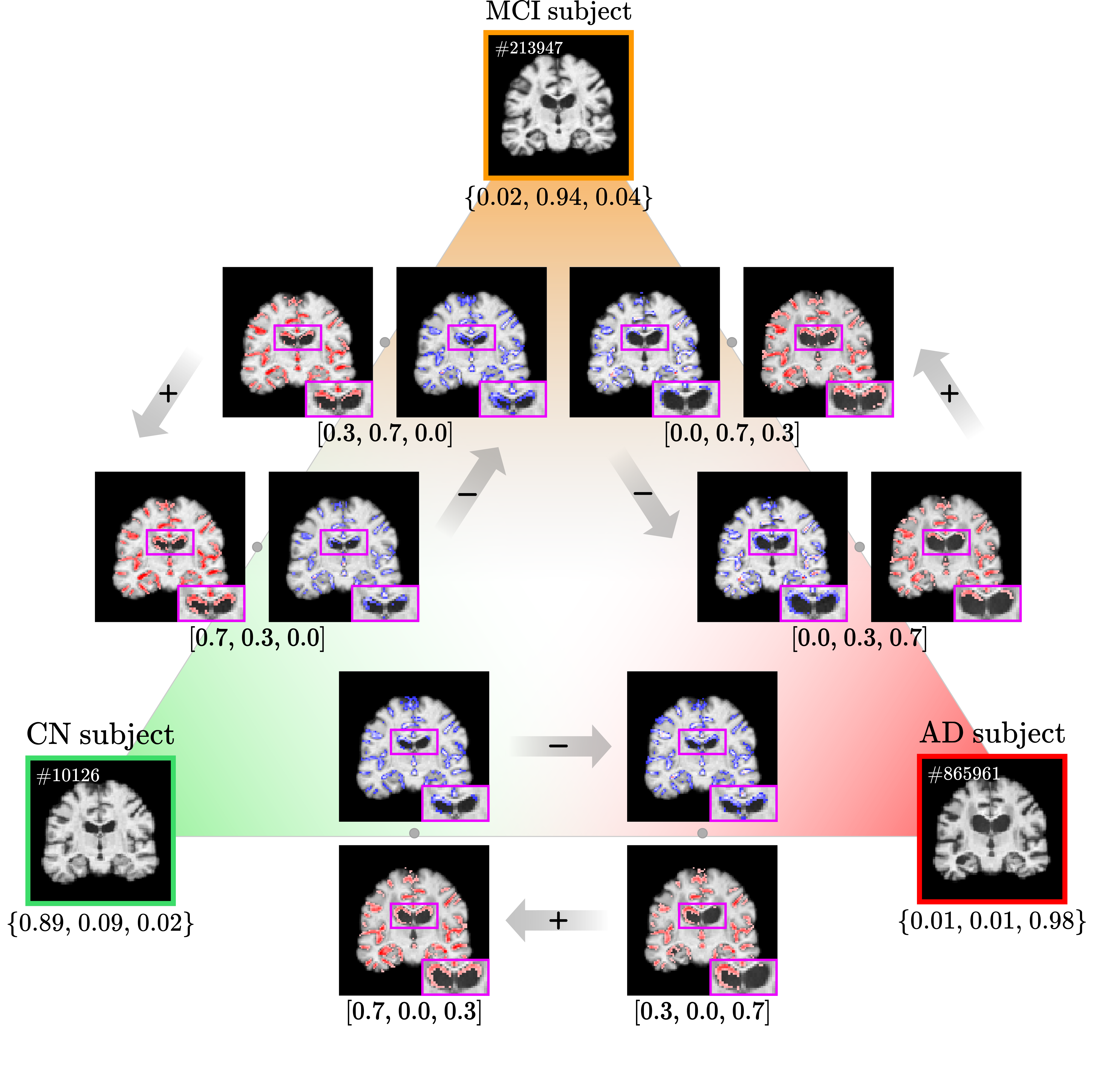}
    \caption{Example of counterfactual map conditioned on interpolated target labels (Subject ID 123\_S\_0106, Image ID on top left corner). The purple boxes correspond to the ventricular region. Parentheses $\{\cdot\}$ and $[\cdot]$ for condition indicate the posterior probability and a target condition, respectively. The +/- signs above the gray arrows denote, respectively, NCC(+) and NCC(-).}
    \label{fig:brain_interpolation}
\end{figure}

\vspace{5pt}\noindent{\bfseries Results and Analysis}

    \noindent In order for qualitative and quantitative evaluation with regard to the visual explanation, we used the longitudinal samples of 12 subjects in ADNI-1/-2, from which the ground-truth maps were created to indicate morphological changes in sMRI according to changes in clinical diagnosis. Details about the longitudinal samples and creating ground-truth maps are given {\supp{in Supplementary S2.2.}}
    It should be noted that none of these images shown there were used in any of our model training procedures.
    
    \noindent{\bfseries (\textit{AD$\rightarrow$CN Counterfactual Maps})}
    For visual explanation of a trained three-class diagnostic model, we applied our CMG and other comparative methods in the literature. Fig. \ref{fig:Brain_result} illustrates their respective results to explain why the input image was diagnosed into AD, instead of CN.
    Notably, our proposed CMG showed the best matching result to the ground-truth map by detecting and highlighting the ventricle enlargement and cortical atrophies. These visual explanations are consistent with the existing clinical neuroscience studies \cite{dickerson2001mri,fan2008spatial,gerardin2009multidimensional}.
    
    The CF map generated by LRP-Z~\cite{bach2015pixel} and DeepLIFT~\cite{shrikumar2017learning} does not clearly show the class discriminative regions. We observe that these approaches focus only on the left hippocampus area while ignoring the right hippocampus area (orange box). Unlike other gradient-based approaches, Guided backpropagation~\cite{selvaraju2017grad}, Integrated gradients~\cite{sundararajan2017axiomatic}, and DeepTaylor~\cite{montavon2017explaining} methods showed some traces of counterfactual reasoning across the image, but unnecessary attributions were observed at the edge or morphological boundaries of the brain. Even though Grad-CAM~\cite{selvaraju2017grad} has shown the class-discriminative visualization, this result slightly captures the coarse regions.
    
    \ks{GAN-based models (\eg, ours and VAGAN~\cite{baumgartner2018visual}) achieve superior results in comparison to other visual explanation methods. However, VAGAN is only successful in mimicking the hypertrophy in the hippocampus regions while failing to capture the increased cortical thickness (in fact, it decreased the cortical thickness in blue-colored regions). In contrast, our method captures almost every subtle region where the cortical thickness was increased while successfully capturing the reduced ventricular and the hypertrophy in the hippocampus. Thus, our CMG module is able to visually explain class-discriminative and fine-grained regions of the brain.}
    
    \vspace{5pt}\noindent{\bfseries (\textit{CN↔MCI↔AD Counterfactual Maps})}
    \ks{In addition to our CMG's ability to produce high-quality CF maps, it is also capable of generating counterfactual explanation maps with regard to diverse conditions in the AD spectrum, which, to the best of our knowledge, cannot be done by the existing comparable methods. 
    We generated \textit{multi-way} CF maps with interpolation-based target conditions setting among the three classes of CN, MCI, and AD and illustrated the results in Fig.~\ref{fig:brain_interpolation}. 
    In this figure, using longitudinal samples of a subject (Subject ID 123\_S\_0106) who had experienced all the clinical stages of CN, MCI, and AD over several years, 
    we produced CF maps under various target conditions. 
    For example, a target condition of $\tilde{\tar}=[0.3,0.7,0]$, where each element accounts for the probability of belonging to the CN, MCI, and AD group, respectively, was used to transform an MCI image $\x_{\text{MCI}}$ to a prodromal CN-like image using the CF map $\M_{\x_{\text{MCI}},\tilde{\tar}}$ (first image in the top row of Fig.~\ref{fig:brain_interpolation}), and the same target condition could be used to transform a CN image $\x_{\text{CN}}$ to a prodromal MCI-like image using the CF map $\M_{\x_{\text{CN}},\tilde{\tar}}$ (second image in the top row).
    }
    
    \ks{Although linearly interpolating between stages of a disease is not a pathologically sound procedure for analyzing the intermediate stages of a disease, our CMG produces sub-optimal\footnote[3]{It is sub-optimal because the progression of disease is not a linear process, but our target condition is.} CF maps in that biomarkers gradually increase or decrease with regards to a given target condition (results on interpolation in finer steps are {\supp{in Supplementary S4}}).
    For example, the size of the ventricle (purple boxes) gradually reduces when interpolating from $\x_{\text{MCI}}$ to $\x_{\text{CN}}$ (\ie, the ``+'' direction), and gradually enlarges when interpolating in the opposite direction (\ie, the ``-'' direction).
    Additionally, we can see that the magnitude of the CF maps $\M_{\x_{\text{CN}}, \tar_{\text{MCI}}}$ (for transforming $\x_{\text{CN}}$ to $\x_{\text{MCI}}$) and $\M_{\x_{\text{MCI}}, \tar_{\text{AD}}}$
    approximately add up to the CF map $\M_{\x_{\text{CN}}, \tar_{\text{AD}}}$ (\ie, $\M_{\x_{\text{CN}}, \tar_{\text{AD}}}\approx\M_{\x_{\text{CN}}, \tar_{\text{MCI}}}+\M_{\x_{\text{MCI}}, \tar_{\text{AD}}}$).
    This indicates that our CMG is able to successfully capture the two extreme tails of most normal brain and most AD-like brain, which strengthens our motivation for using these maps as the source of attention in the XGA module (see Section~\ref{sec:GMG}).  
    }
    
    \vspace{5pt}\noindent{\bfseries (Quantitative Evaluation)} To quantitatively assess the quality of our generated CF maps, we calculated the normalized cross-correlation (NCC) score between generated CF maps and ground-truth maps as follows \cite{baumgartner2018visual}. The NCC score measures the similarity between two samples in a normalized setting where higher NCC scores denote higher similarity. Thus, NCC can be helpful when two samples have a different magnitude of signals. 
    Here, we denote the ground-truth maps and CF maps for transforming CN←MCI, MCI←AD, CN←AD as the ``+'' direction and CN→MCI, MCI→AD, CN→AD as the ``-'' direction (see Fig.~\ref{fig:brain_interpolation}), and calculate NCC(+) and NCC(-) for each.
    
    \ks{The scores for LRP-Z~\cite{bach2015pixel} and DeepLIFT~\cite{shrikumar2017learning} are understandably low because they can only capture the least number of class-discriminative features as seen in Fig.~\ref{fig:Brain_result}. Integrated Gradients~\cite{selvaraju2017grad} can capture the class-discriminative features in a group-wise manner, \ie, the values of their CF maps do not differ significantly for different subjects, and so their NCC score, which is a subject-wise correlation score, is very low. Guided Backprop~\cite{selvaraju2017grad}, DeepTaylor~\cite{montavon2017explaining}, and GradCAM~\cite{selvaraju2017grad} can capture some class-discriminative features, but only in a coarse-grained manner.}
    We found that VAGAN~\cite{baumgartner2018visual} has captured some meaningful regions for disease localization. However, NCC scores of our proposed CMG are higher than VAGAN (Table~\ref{tab:compared_exp}) because our CF maps can localize biomarkers throughout the brain, while the CF maps of VAGAN fail to capture the class-discriminative features in the cortical regions. Unlike the competing methods, which are built on top of binary classifiers, our LEAR framework can fully utilize various backbone diagnostic models (\eg, ResNet18, VoxCNN, and SonoNet16) for binary and multi-class classification tasks. The full and comprehensive results are provided {\supp{in Supplementary S3.1}}.
    
    \ks{One interesting phenomenon across methods is the lower NCC scores in the ``-'' direction, \ie, NCC(-). 
    A simple hypothesis we made was that more (difficult) processes are required for subtracting, which happens mostly in the \mbox{``-''} direction, than for adding certain regions of a brain.
    For example, a baseline CN image tends to have more gray matter (\ie, gray colored tissues) in certain biomarker regions than its progressed AD image. These gray matter regions also tend to contain more complex morphological features than other (\ie, white matter and cerebrospinal fluid) regions, which makes transforming to gray matter (\ie, a ``-'' operation) more difficult given that morphological features need to be drawn on top of those regions.
    With that said, our LEAR framework could mitigate this gap between NCC(+) and NCC(-) using the guidance map $\M_\text{guide}$ which allows our framework to account for the both extreme cases of most normal and most AD-like brains.}

\begin{table}[t]\scriptsize \setlength{\tabcolsep}{1.pt}
    \caption{Normalized Cross-Correlation (NCC) scores with comparison methods on the ADNI dataset. We differentiated NCC scores for each generation direction of the counterfactual map. The +/- signs indicate different directions of the counterfactual map (see Fig.~\ref{fig:brain_interpolation}).}
    \centering
    \label{tab:compared_exp}
    \begin{tabular}{ccccccc}
    \toprule
    \multicolumn{1}{c}{\multirow{2}{*}{\textbf{Scenario}}} & \multicolumn{2}{c}{\textbf{CN $\leftrightarrow$ MCI}} & \multicolumn{2}{c}{\textbf{MCI $\leftrightarrow$ AD}} & \multicolumn{2}{c}{\textbf{CN $\leftrightarrow$ AD}}\\
    \cmidrule(lr){2-3} \cmidrule(lr){4-5} \cmidrule(lr){6-7}
    & \textbf{NCC(+)} & \textbf{NCC(-)} & \textbf{NCC(+)} & \textbf{NCC(-)} & \textbf{NCC(+)} & \textbf{NCC(-)} \\

\midrule
    LRP-Z~\cite{bach2015pixel} & \multicolumn{1}{c}{0.005} &\multicolumn{1}{c}{0.005} &\multicolumn{1}{c}{0.006} &\multicolumn{1}{c}{0.004} & \multicolumn{1}{c}{0.008} & \multicolumn{1}{c}{0.005} \\
    Integrated Gradients~\cite{sundararajan2017axiomatic} & \multicolumn{1}{c}{0.006} &\multicolumn{1}{c}{0.007} &\multicolumn{1}{c}{0.007} &\multicolumn{1}{c}{0.007} &\multicolumn{1}{c}{0.006} &\multicolumn{1}{c}{0.005}\\
    DeepLIFT~\cite{shrikumar2017learning} &\multicolumn{1}{c}{0.004} &\multicolumn{1}{c}{0.005} &\multicolumn{1}{c}{0.006} &\multicolumn{1}{c}{0.004} &\multicolumn{1}{c}{0.005} &\multicolumn{1}{c}{0.004}\\
    Guided Backprop~\cite{selvaraju2017grad} &\multicolumn{1}{c}{0.199} &\multicolumn{1}{c}{0.158} &\multicolumn{1}{c}{0.212} &\multicolumn{1}{c}{0.163} &\multicolumn{1}{c}{0.239} &\multicolumn{1}{c}{0.204}\\
    DeepTaylor~\cite{montavon2017explaining} &\multicolumn{1}{c}{0.143} &\multicolumn{1}{c}{0.172} &\multicolumn{1}{c}{0.112} &\multicolumn{1}{c}{0.108} &\multicolumn{1}{c}{0.132} &\multicolumn{1}{c}{0.118}\\
    Grad-CAM~\cite{selvaraju2017grad} &\multicolumn{1}{c}{0.201} &\multicolumn{1}{c}{0.188} &\multicolumn{1}{c}{0.215} &\multicolumn{1}{c}{0.227} &\multicolumn{1}{c}{0.227} &\multicolumn{1}{c}{0.214}\\
    VA-GAN~\cite{baumgartner2018visual} &\multicolumn{1}{c}{0.283} &\multicolumn{1}{c}{0.186} &\multicolumn{1}{c}{0.285} &\multicolumn{1}{c}{0.257} &\multicolumn{1}{c}{0.317} &\multicolumn{1}{c}{0.298}\\
    \midrule
    \textbf{Ours} &\multicolumn{1}{c}{\textbf{0.364}} &\multicolumn{1}{c}{\textbf{0.289}} &\multicolumn{1}{c}{\textbf{0.299}} &\multicolumn{1}{c}{\textbf{0.297}} &\multicolumn{1}{c}{\textbf{0.366}} &\multicolumn{1}{c}{\textbf{0.312}}\\
\bottomrule
\end{tabular}
\end{table}

\subsection{Diagnostic Model Reinforcement}
In this section, we demonstrate the effectiveness of our LEAR framework in reinforcing diagnostic models. To do so, we have divided this section into three parts. First, we consider the CF map transformation, \ie, Eq.~(\ref{cm_map_original}), as a baseline method for our work. Second, we compare our LEAR framework with state-of-the-art attention methods. Third, we demonstrate the effectiveness of the optimized CF map transformation, \ie, Eq.~(\ref{cm_renew}), in improving the quality of visual explanation as well as the performance of diagnostic models.

To verify the effectiveness of our proposed XGA module and its produced guidance map, we compare the diagnostic performance in accuracy (ACC) and multi-class area under the receiver operating characteristic curve (mAUC).
\ks{Note that we use a five-fold cross-validation setting for all experiments, and use the same indices for all the comparison methods with no data leakage.}

\subsubsection{Reinforcement via Augmentation}
As a baseline method for diagnostic model reinforcement, we utilize the CF map without XGA injection (\ie, Eq.~(\ref{cm_map_original})) to produce synthesized images to \ks{augment training samples and use those to update the backbone diagnostic model.}
Specifically, using a CF map defined by Eq.~(\ref{cm_map_original}), we transformed all train data samples with target labels other than their ground-truth label. For three-class experiments, we produced transformed images with two other target labels. For example, if the input image is an AD subject, we produced NC-transformed and MCI-transformed images for that input image.
\ks{Finally, those transformed images were used to fine-tune the backbone diagnostic models.
We decided to use this augmentation method as a baseline method for our work because it is one of the simplest ways to utilize the CF map in reinforcing the diagnostic performance.
To this end, we report the comparison between the backbone, baseline, and our method in Table~\ref{tab:synthesis}.
The improvement (+3.7\%) in the classification accuracy of our method over that of the baseline method suggests that our CF maps can indeed capture class-discriminative information and also indicates that these kinds of visual explanation can guide and reinforce a diagnostic model, which supports the motivation behind this study.}
In the following paragraphs, we will demonstrate that the LEAR framework can further reinforce the diagnostic models with guidance from visual explanation using CF maps.

\begin{table}[t]\scriptsize \setlength{\tabcolsep}{4pt}
    \centering
    \caption{Comparison of performance (ACC) among the backbone, augmentation, and the attention with guidance on ADNI dataset.}
    \label{tab:synthesis}
    \begin{tabular}{cccc}
    \toprule
   \multicolumn{1}{c}{\multirow{2}{*}{\textbf{Setting}}} & \multicolumn{3}{c}{\textbf{ResNet18}}\\
    \cmidrule(lr){2-4}
    & \textbf{backbone} & \textbf{augmentation} & \textbf{ours}\\
\midrule
    CN \textit{vs.} MCI \textit{vs.} AD&\multicolumn{1}{c}{0.5802} &\multicolumn{1}{c}{0.5883}&\multicolumn{1}{c}{\textbf{0.6715}}\\
    CN \textit{vs.} MCI &\multicolumn{1}{c}{0.6479} &\multicolumn{1}{c}{0.6856} &\multicolumn{1}{c}{\textbf{0.7436}}\\
    sMCI \textit{vs.} pMCI & \multicolumn{1}{c}{0.6946} & \multicolumn{1}{c}{0.7162} &\multicolumn{1}{c}{\textbf{0.7703}}\\
    MCI \textit{vs.} AD &\multicolumn{1}{c}{0.7965} &\multicolumn{1}{c}{0.8333}&\multicolumn{1}{c}{\textbf{0.8716}}\\
    CN \textit{vs.} AD & \multicolumn{1}{c}{0.8898} & \multicolumn{1}{c}{0.9231}&\multicolumn{1}{c}{\textbf{0.9489}}\\
\bottomrule
\end{tabular}
\end{table}

\subsubsection{Comparison to Other Diagnostic Models}
To demonstrate the ability of our LEAR framework in reinforcing diagnostic models, we have pre-trained and fixed the weights of a three-class backbone ResNet18 diagnostic model and re-implemented state-of-the-art diagnostic models. To this end, we compare the performances of state-of-the-art \textit{attention-guided} models and \textit{conventional} CNN models in Table~\ref{tab:3class_cls}.

Notably, our work demonstrates significant improvement over the ResNet18 backbone model (ACC +15.74\%) as well as the state-of-the-art CNN models (mean ACC +13.64\%).
In comparison to conventional CNNs, attention-guided diagnostic models (\eg, Li~\etal~\cite{li2019novel} and Lian~\etal~\cite{lian2020attention}) mostly excel in diagnostic performances. However, these models are guided by \textit{conventional} visual attribution methods, such as CAM, that can only provide coarse-grained guidance. Our LEAR framework, synthesizing and exploiting fine-grained guidance, outperformed all the competing methods by large margins in mAUC and ACC. It is noteworthy that the performance improvements were obtained for all the diagnostic models considered in our experiments (\ie, ResNet18, VoxCNN, and SonoNet16) in the experiments equally.
\ks{Furthermore, as most of the comparing works were proposed as binary diagnostic models, we have performed a comprehensive binary diagnosis comparison and presented {\supp{in Supplementary S3.2}}.
Our LEAR framework outperforms all comparing models in all binary class settings (mean ACC: CN $vs.$ MCI +14.80\%, sMCI $vs.$ pMCI +10.82\%, MCI $vs.$ AD +12.83\%, CN $vs.$ AD +7.69\%).}

\begin{table}[t]\scriptsize \setlength{\tabcolsep}{3.pt}
    \centering
    \caption{Comparison of performance on the multi-class (\ie, CN \textit{vs.} MCI \textit{vs.} AD) classification scenario on the ADNI dataset.}
    \label{tab:3class_cls}
    \begin{tabular}{cccc}
    \toprule
   \multicolumn{1}{c}{\textbf{Guidance}}&\multicolumn{1}{c}{\textbf{Models}} & \multicolumn{1}{c}{\textbf{mAUC}} & \multicolumn{1}{c}{\textbf{ACC}}\\
\midrule
    & ResNet18~\cite{he2016deep}&\multicolumn{1}{c}{\text{0.7501} $\pm$ \text{0.046}} &\multicolumn{1}{c}{\text{0.5802} $\pm$ \text{0.041}}\\
    &SonoNet16~\cite{baumgartner2017sononet}&\multicolumn{1}{c}{\text{0.7452} $\pm$ \text{0.069}} &\multicolumn{1}{c}{\text{0.5912} $\pm$ \text{0.056}}\\
    &VoxCNN~\cite{korolev2017residual}&\multicolumn{1}{c}{\text{0.7732} $\pm$ \text{0.034}} &\multicolumn{1}{c}{\text{0.5863} $\pm$ \text{0.045}}\\
    &Liu \etal~\cite{liu2020design}&\multicolumn{1}{c}{\text{0.7016} $\pm$ \text{0.056}} &\multicolumn{1}{c}{\text{0.5468} $\pm$ \text{0.069}}\\
    &Jin \etal~\cite{jin2019attention}&\multicolumn{1}{c}{\text{0.7294} $\pm$ \text{0.055}} &\multicolumn{1}{c}{\text{0.5901} $\pm$ \text{0.041}}\\
    \midrule
    \hspace{0.cm}\parbox[t]{2mm}{\multirow{3}{*}{\rotatebox[origin=c]{0}{$\checkmark$}}}&Li \etal~\cite{li2019novel}&\multicolumn{1}{c}{\text{0.7559} $\pm$ \text{0.038}} &\multicolumn{1}{c}{\text{0.6115} $\pm$ \text{0.062}}\\
    &Lian \etal~\cite{lian2020attention}&\multicolumn{1}{c}{\text{0.7671} $\pm$ \text{0.075}} &\multicolumn{1}{c}{\text{0.6257} $\pm$ \text{0.059}}\\
    &\textbf{Ours (ResNet18 + XGA)}&\multicolumn{1}{c}{\textbf{0.8123} $\pm$ \textbf{0.052}} &\multicolumn{1}{c}{\textbf{0.6715} $\pm$ \textbf{0.051}}\\

\bottomrule
\end{tabular}
\end{table}

\begin{figure}[t]
    \centering
    \includegraphics[width=1.\linewidth]{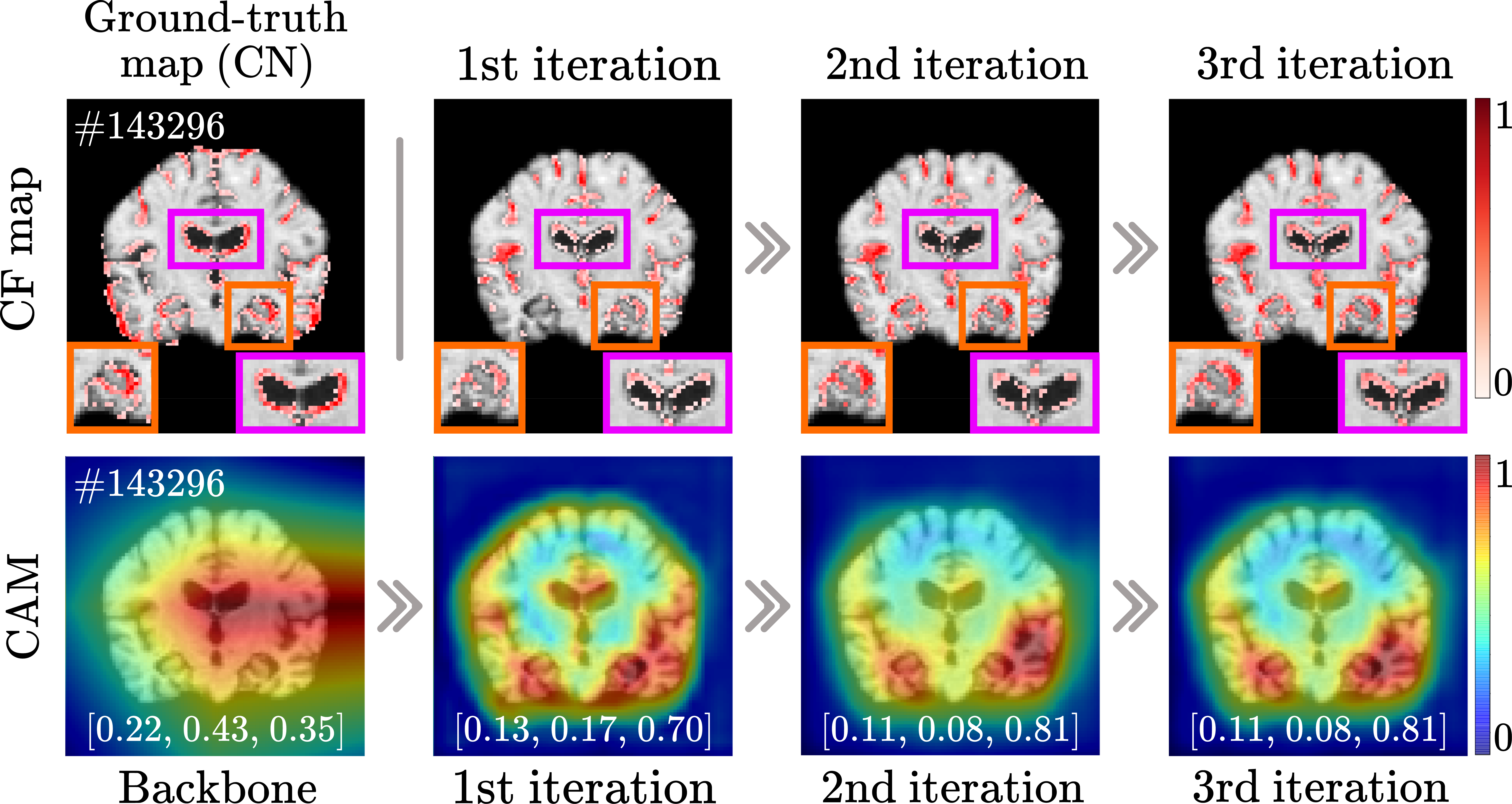}
    \caption{Counterfactual map and CAM visualization of XGA-injected ResNet18 on the CN \textit{vs.} MCI \textit{vs.} AD scenario with self-iterative training. The values at the bottom of brain images (Subject ID 005\_S\_0223) are the model's softmax activated logits.}\label{fig:cam_visual}
\end{figure}

\subsection{Iterative Explanation-Reinforcement Learning}
Here, we demonstrate how the iterative learning scheme of our LEAR framework can further improve the diagnostic performances and, thereby, the quality of visual explanation. 

\subsubsection{Effects in Generalization of a Diagnostic Model}
\ks{We applied three iterations of our LEAR framework and presented the results in Table~\ref{table:self_iter}.
In comparison to the backbone model, the iterations of our LEAR framework have increased the accuracy by +5.49\%, +10.64\%, and +10.50\%, respectively, for each iteration.}

For a visual inspection of the changes in class-relevant feature representations, in Fig. \ref{fig:cam_visual}, we present the CAM visualization along with the CF map over three iterations of EU and RU learning for an AD sample (Subject ID 005\_S\_0223), which our backbone ResNet18 diagnostic model misclassified with a low class probability.
\ks{As shown in Fig.~\ref{fig:cam_visual}, the XGA module excels in cases where the confidence of the predictive probability (\ie, values at the bottom of each image) of the backbone diagnostic model is low. Specifically, the CAM obtained from the backbone network vaguely highlights the ventricular region (\ie, the center of an image), whereas the CAM results after iterative learning focus on more meaningful and fine-grained regions of the cortex and hippocampus. Likewise, the first iteration CF map neglected to highlight the hippocampus region (orange box) associated with the AD progression, but the second and third iteration CF maps clearly observed the hypertrophy in the hippocampus region with an equivalent intensity. Thus, the XGA module of our LEAR framework improves the diagnostic model not only in terms of performance but also in terms of innate interpretability of a diagnostic model.} 
\ks{Interestingly, we found that diagnostic scores converge after second iteration, but the \ks{CAM results} continue to qualitatively improve even after the second iteration. The more the iterations that were performed, the more fine-grained and class-discriminative the CAM results became.}

\subsubsection{Effects in Visual Explanation}
\begin{table}[t]\scriptsize \setlength{\tabcolsep}{4.pt}
    \centering
    \caption{Comparison of performance (ACC) among various iterations on the ADNI dataset.}
    \label{table:self_iter}
    \begin{tabular}{ccccc}
    \toprule
   \multicolumn{1}{c}{\multirow{2}{*}{\textbf{Setting}}} & \multicolumn{4}{c}{\textbf{ResNet18}}\\
    \cmidrule(lr){2-5}
    & \textbf{Backbone} & \textbf{1st} & \textbf{2nd} & \textbf{3rd}\\
\midrule
    CN \textit{vs.} MCI \textit{vs.} AD &\multicolumn{1}{c}{0.5802}&\multicolumn{1}{c}{0.6347} &\multicolumn{1}{c}{0.6715} &\multicolumn{1}{c}{0.6715}\\
    CN \textit{vs.} MCI &\multicolumn{1}{c}{0.6479} &\multicolumn{1}{c}{0.7014} &\multicolumn{1}{c}{{0.7436}} &\multicolumn{1}{c}{0.7436}\\
    sMCI \textit{vs.} pMCI &\multicolumn{1}{c}{0.6946}&\multicolumn{1}{c}{0.7381} &\multicolumn{1}{c}{{0.7703}} &\multicolumn{1}{c}{0.7703}\\
    MCI \textit{vs.} AD &\multicolumn{1}{c}{0.7965}&\multicolumn{1}{c}{0.8396} &\multicolumn{1}{c}{{0.8716}} &\multicolumn{1}{c}{0.8716}\\
    CN \textit{vs.} AD &\multicolumn{1}{c}{0.8898} & \multicolumn{1}{c}{0.9229} & \multicolumn{1}{c}{{0.9515}} &\multicolumn{1}{c}{0.9489}\\
\bottomrule
\end{tabular}
\end{table}

\begin{figure}[t]
    \centering
    \includegraphics[width=.95\linewidth]{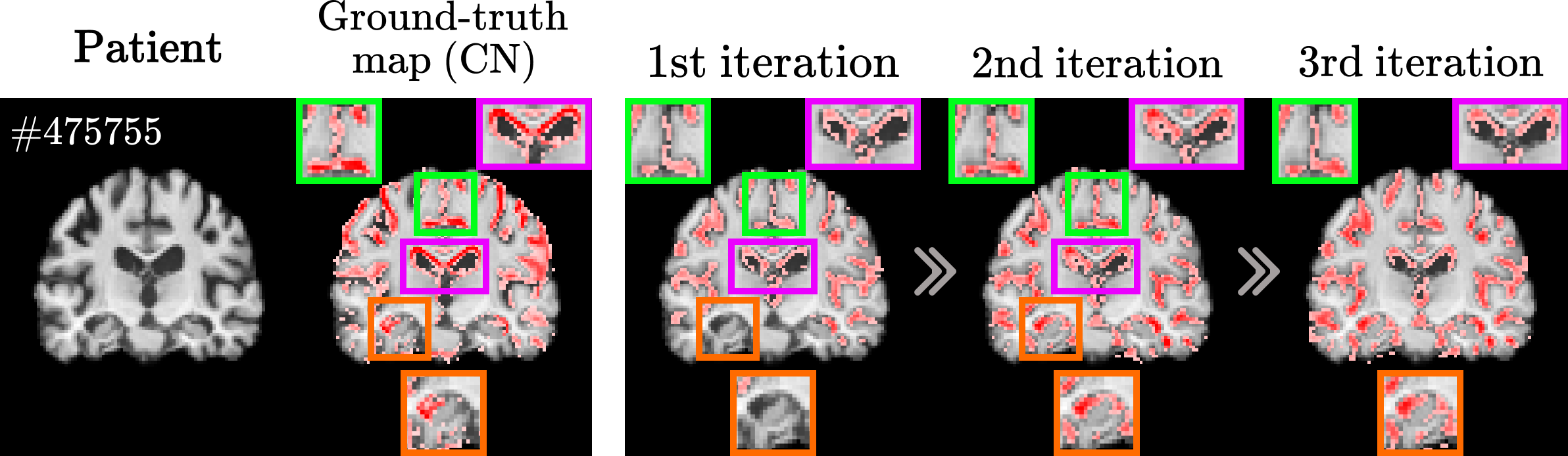}
    \caption{Reinforced counterfactual map visualization by using iterative optimization on trained ResNet18 (Subject ID 131\_S\_0123, Image ID on top left corner). The purple, orange, and green boxes correspond to the ventricular, hippocampus, and cortex regions, respectively.}\label{fig:cf_visual}
\end{figure}

\ks{\noindent First, we selected a CN and AD image from a subject (Subject ID 131\_S\_0123) whose CF map was of unsatisfying quality.
Then, we produced the ground-truth map for CN←AD transformation and CF maps for three iterations of our LEAR framework (Fig.~\ref{fig:cf_visual}).
Note that the first iteration CF map does not benefit from reinforcement because the CF map is defined by Eq.~(\ref{cm_map_original}) at this iteration and is redefined by Eq.~(\ref{cm_renew}) from the second iteration onwards.}

\ks{In the first-iteration CF map, the attribution completely ignores the hypertrophy in the hippocampus (orange box). The attributions in the cortical (green box) and ventricular (purple box) regions are also weak and noisy, making the visual explanation pathologically unreliable.
However, the second-iteration CF map successfully captures the hypertrophy in the hippocampus and the attribution to the cortex regions is clearer, but the attribution in the ventricles has become nosier. Finally, the third-iteration CF map clears up the noisy attribution in the ventricles. More diverse results are presented in \supp{Supplementary S6}.
}

\section{Conclusion}\label{sec5}
    In this work, we proposed an approach for producing high-quality visual explanations in hypothetical scenarios through counterfactual reasoning maps as well as reinforcing diagnostic performances. Our LEAR framework achieved state-of-the-art performances qualitatively and quantitatively upon the ADNI dataset.

\section*{Acknowledgement}
This study was supported by the Institute of Information \& Communications Technology Planning \& Evaluation (IITP) grant funded by the Korea Government (MSIT) No. 2017-0-01779 (A machine learning and statistical inference framework for explainable artificial intelligence) and No. 2019-0-00079 (Department of Artificial Intelligence (Korea University)). 

\bibliographystyle{IEEEtran}
\bibliography{main}

\begin{thebibliography}{10}
\providecommand{\url}[1]{#1}
\csname url@samestyle\endcsname
\providecommand{\newblock}{\relax}
\providecommand{\bibinfo}[2]{#2}
\providecommand{\BIBentrySTDinterwordspacing}{\spaceskip=0pt\relax}
\providecommand{\BIBentryALTinterwordstretchfactor}{4}
\providecommand{\BIBentryALTinterwordspacing}{\spaceskip=\fontdimen2\font plus
\BIBentryALTinterwordstretchfactor\fontdimen3\font minus
  \fontdimen4\font\relax}
\providecommand{\BIBforeignlanguage}[2]{{%
\expandafter\ifx\csname l@#1\endcsname\relax
\typeout{** WARNING: IEEEtran.bst: No hyphenation pattern has been}%
\typeout{** loaded for the language `#1'. Using the pattern for}%
\typeout{** the default language instead.}%
\else
\language=\csname l@#1\endcsname
\fi
#2}}
\providecommand{\BIBdecl}{\relax}
\BIBdecl

\bibitem{alzheimer20192019}
A.~Association \emph{et~al.}, ``2019 {A}lzheimer's disease facts and figures,''
  \emph{Alzheimer's \& dementia}, vol.~15, no.~3, pp. 321--387, 2019.

\bibitem{alzheimer20182018}
A.~Association, ``2018 {A}lzheimer's disease facts and figures,''
  \emph{Alzheimer's \& Dementia}, vol.~14, no.~3, pp. 367--429, 2018.

\bibitem{li2013variation}
S.~Li, O.~Okonkwo, M.~Albert, and M.-C. Wang, ``Variation in variables that
  predict progression from {MCI} to {AD} dementia over duration of follow-up,''
  \emph{American Journal of Alzheimer's Disease (Columbia, Mo.)}, vol.~2,
  no.~1, p.~12, 2013.

\bibitem{frisoni2010clinical}
G.~B. Frisoni, N.~C. Fox, C.~R. Jack, P.~Scheltens, and P.~M. Thompson, ``The
  clinical use of structural {MRI} in {A}lzheimer disease,'' \emph{Nature
  Reviews Neurology}, vol.~6, no.~2, pp. 67--77, 2010.

\bibitem{kloppel2008automatic}
S.~Kl{\"o}ppel, C.~M. Stonnington, C.~Chu, B.~Draganski, R.~I. Scahill, J.~D.
  Rohrer, N.~C. Fox, C.~R. Jack~Jr, J.~Ashburner, and R.~S. Frackowiak,
  ``Automatic classification of {MR} scans in {A}lzheimer's disease,''
  \emph{Brain}, vol. 131, no.~3, pp. 681--689, 2008.

\bibitem{hinrichs2009spatially}
C.~Hinrichs, V.~Singh, L.~Mukherjee, G.~Xu, M.~K. Chung, S.~C. Johnson,
  A.~D.~N. Initiative \emph{et~al.}, ``Spatially augmented {LP}boosting for
  {AD} classification with evaluations on the {ADNI} dataset,''
  \emph{NeuroImage}, vol.~48, no.~1, pp. 138--149, 2009.

\bibitem{zhang2011multimodal}
D.~Zhang, Y.~Wang, L.~Zhou, H.~Yuan, D.~Shen, A.~D.~N. Initiative
  \emph{et~al.}, ``Multimodal classification of {A}lzheimer's disease and mild
  cognitive impairment,'' \emph{NeuroImage}, vol.~55, no.~3, pp. 856--867,
  2011.

\bibitem{suk2013deep}
H.-I. Suk and D.~Shen, ``Deep learning-based feature representation for
  {AD}/{MCI} classification,'' in \emph{International Conference on Medical
  Image Computing and Computer-Assisted Intervention}.\hskip 1em plus 0.5em
  minus 0.4em\relax Springer, 2013, pp. 583--590.

\bibitem{korolev2017residual}
S.~Korolev, A.~Safiullin, M.~Belyaev, and Y.~Dodonova, ``Residual and plain
  convolutional neural networks for {3D} brain {MRI} classification,'' in
  \emph{IEEE 14th International Symposium on Biomedical Imaging}.\hskip 1em
  plus 0.5em minus 0.4em\relax IEEE, 2017, pp. 835--838.

\bibitem{suk2017deep}
H.-I. Suk, S.-W. Lee, D.~Shen, A.~D.~N. Initiative \emph{et~al.}, ``Deep
  ensemble learning of sparse regression models for brain disease diagnosis,''
  \emph{Medical Image Analysis}, vol.~37, pp. 101--113, 2017.

\bibitem{arrieta2020explainable}
A.~B. Arrieta, N.~D{\'\i}az-Rodr{\'\i}guez, J.~Del~Ser, A.~Bennetot, S.~Tabik,
  A.~Barbado, S.~Garc{\'\i}a, S.~Gil-L{\'o}pez, D.~Molina, R.~Benjamins
  \emph{et~al.}, ``Explainable artificial intelligence ({XAI}): Concepts,
  taxonomies, opportunities and challenges toward responsible {AI},''
  \emph{Information Fusion}, vol.~58, pp. 82--115, 2020.

\bibitem{singh2020explainable}
A.~Singh, S.~Sengupta, and V.~Lakshminarayanan, ``Explainable deep learning
  models in medical image analysis,'' \emph{Journal of Imaging}, vol.~6, no.~6,
  p.~52, 2020.

\bibitem{gunning2019xai}
D.~Gunning, M.~Stefik, J.~Choi, T.~Miller, S.~Stumpf, and G.-Z. Yang,
  ``{XAI}—explainable artificial intelligence,'' \emph{Science Robotics},
  vol.~4, no.~37, 2019.

\bibitem{mascharka2018transparency}
D.~Mascharka, P.~Tran, R.~Soklaski, and A.~Majumdar, ``Transparency by design:
  Closing the gap between performance and interpretability in visual
  reasoning,'' in \emph{IEEE Conference on Computer Vision and Pattern
  Recognition}, 2018, pp. 4942--4950.

\bibitem{rudin2019stop}
C.~Rudin, ``Stop explaining black box machine learning models for high stakes
  decisions and use interpretable models instead,'' \emph{Nature Machine
  Intelligence}, vol.~1, no.~5, pp. 206--215, 2019.

\bibitem{gilpin2018explaining}
L.~H. Gilpin, D.~Bau, B.~Z. Yuan, A.~Bajwa, M.~Specter, and L.~Kagal,
  ``Explaining explanations: An overview of interpretability of machine
  learning,'' in \emph{IEEE International Conference on Data Science and
  Advanced Analytics}.\hskip 1em plus 0.5em minus 0.4em\relax IEEE, 2018, pp.
  80--89.

\bibitem{selvaraju2017grad}
R.~R. Selvaraju, M.~Cogswell, A.~Das, R.~Vedantam, D.~Parikh, and D.~Batra,
  ``Grad-{CAM}: Visual explanations from deep networks via gradient-based
  localization,'' in \emph{IEEE International Conference on Computer Vision},
  2017, pp. 618--626.

\bibitem{sundararajan2017axiomatic}
M.~Sundararajan, A.~Taly, and Q.~Yan, ``Axiomatic attribution for deep
  networks,'' in \emph{34th International Conference on Machine Learning},
  2017, pp. 3319--3328.

\bibitem{bach2015pixel}
S.~Bach, A.~Binder, G.~Montavon, F.~Klauschen, K.-R. M{\"u}ller, and W.~Samek,
  ``On pixel-wise explanations for non-linear classifier decisions by
  layer-wise relevance propagation,'' \emph{Public Library of Science ONE},
  vol.~10, no.~7, p. e0130140, 2015.

\bibitem{montavon2017explaining}
G.~Montavon, S.~Lapuschkin, A.~Binder, W.~Samek, and K.-R. M{\"u}ller,
  ``Explaining nonlinear classification decisions with deep taylor
  decomposition,'' \emph{Pattern Recognition}, vol.~65, pp. 211--222, 2017.

\bibitem{shrikumar2017learning}
A.~Shrikumar, P.~Greenside, and A.~Kundaje, ``Learning important features
  through propagating activation differences,'' in \emph{34th International
  Conference on Machine Learning}, 2017, pp. 3145--3153.

\bibitem{zeiler2014visualizing}
M.~D. Zeiler and R.~Fergus, ``Visualizing and understanding convolutional
  networks,'' in \emph{European Conference on Computer Vision}.\hskip 1em plus
  0.5em minus 0.4em\relax Springer, 2014, pp. 818--833.

\bibitem{goyal2019counterfactual}
Y.~Goyal, Z.~Wu, J.~Ernst, D.~Batra, D.~Parikh, and S.~Lee, ``Counterfactual
  visual explanations,'' in \emph{36th International Conference on Machine
  Learning}, 2019, pp. 2376--2384.

\bibitem{goyal2019explaining}
Y.~Goyal, A.~Feder, U.~Shalit, and B.~Kim, ``Explaining classifiers with causal
  concept effect ({C}a{CE}),'' \emph{arXiv preprint arXiv:1907.07165}, 2019.

\bibitem{wang2020scout}
P.~Wang and N.~Vasconcelos, ``{SCOUT}: Self-aware discriminant counterfactual
  explanations,'' in \emph{IEEE/CVF Conference on Computer Vision and Pattern
  Recognition}, 2020, pp. 8981--8990.

\bibitem{wachter2017counterfactual}
S.~Wachter, B.~Mittelstadt, and C.~Russell, ``Counterfactual explanations
  without opening the black box: Automated decisions and the {GDPR},''
  \emph{Harv. JL \& Tech.}, vol.~31, p. 841, 2017.

\bibitem{baumgartner2018visual}
C.~F. Baumgartner, L.~M. Koch, K.~Can~Tezcan, J.~Xi~Ang, and E.~Konukoglu,
  ``Visual feature attribution using wasserstein {GAN}s,'' in \emph{IEEE
  Conference on Computer Vision and Pattern Recognition}, 2018, pp. 8309--8319.

\bibitem{bass2020icam}
C.~Bass, M.~da~Silva, C.~Sudre, P.-D. Tudosiu, S.~Smith, and E.~Robinson,
  ``{ICAM}: Interpretable classification via disentangled representations and
  feature attribution mapping,'' in \emph{Advances in Neural Information
  Processing Systems}, vol.~33, 2020.

\bibitem{chang2018explaining}
C.-H. Chang, E.~Creager, A.~Goldenberg, and D.~Duvenaud, ``Explaining image
  classifiers by counterfactual generation,'' in \emph{International Conference
  on Learning Representations}, 2019.

\bibitem{van2019interpretable}
A.~Van~Looveren and J.~Klaise, ``Interpretable counterfactual explanations
  guided by prototypes,'' \emph{arXiv preprint arXiv:1907.02584}, 2019.

\bibitem{sauer2021counterfactual}
A.~Sauer and A.~Geiger, ``Counterfactual generative networks,'' in
  \emph{International Conference on Learning Representations}, 2021.

\bibitem{dash2020counterfactual}
S.~Dash and A.~Sharma, ``Counterfactual generation and fairness evaluation
  using adversarially learned inference,'' \emph{arXiv preprint
  arXiv:2009.08270}, 2020.

\bibitem{liu2020design}
S.~Liu, C.~Yadav, C.~Fernandez-Granda, and N.~Razavian, ``On the design of
  convolutional neural networks for automatic detection of {A}lzheimer’s
  disease,'' in \emph{Machine Learning for Health Workshop}.\hskip 1em plus
  0.5em minus 0.4em\relax PMLR, 2020, pp. 184--201.

\bibitem{jin2019attention}
D.~Jin, J.~Xu, K.~Zhao, F.~Hu, Z.~Yang, B.~Liu, T.~Jiang, and Y.~Liu,
  ``Attention-based {3D} convolutional network for {A}lzheimer’s disease
  diagnosis and biomarkers exploration,'' in \emph{IEEE 16th International
  Symposium on Biomedical Imaging}.\hskip 1em plus 0.5em minus 0.4em\relax
  IEEE, 2019, pp. 1047--1051.

\bibitem{zhang2021explainable}
X.~Zhang, L.~Han, W.~Zhu, L.~Sun, and D.~Zhang, ``An explainable {3D} residual
  self-attention deep neural network for joint atrophy localization and
  {A}lzheimer's disease diagnosis using structural {MRI},'' \emph{IEEE Journal
  of Biomedical and Health Informatics}, 2021.

\bibitem{lian2020attention}
C.~Lian, M.~Liu, Y.~Pan, and D.~Shen, ``Attention-guided hybrid network for
  dementia diagnosis with structural {MR} images,'' \emph{IEEE Transactions on
  Cybernetics}, 2020.

\bibitem{zhou2016learning}
B.~Zhou, A.~Khosla, A.~Lapedriza, A.~Oliva, and A.~Torralba, ``Learning deep
  features for discriminative localization,'' in \emph{IEEE Conference on
  Computer Vision and Pattern Recognition}, 2016, pp. 2921--2929.

\bibitem{li2019novel}
Q.~Li, X.~Xing, Y.~Sun, B.~Xiao, H.~Wei, Q.~Huo, M.~Zhang, X.~S. Zhou, Y.~Zhan,
  Z.~Xue \emph{et~al.}, ``Novel iterative attention focusing strategy for joint
  pathology localization and prediction of {MCI} progression,'' in
  \emph{International Conference on Medical Image Computing and
  Computer-Assisted Intervention}.\hskip 1em plus 0.5em minus 0.4em\relax
  Springer, 2019, pp. 307--315.

\bibitem{fong2017interpretable}
R.~C. Fong and A.~Vedaldi, ``Interpretable explanations of black boxes by
  meaningful perturbation,'' in \emph{IEEE International Conference on Computer
  Vision}, 2017, pp. 3429--3437.

\bibitem{dhurandhar2018explanations}
A.~Dhurandhar, P.-Y. Chen, R.~Luss, C.-C. Tu, P.~Ting, K.~Shanmugam, and
  P.~Das, ``Explanations based on the missing: Towards contrastive explanations
  with pertinent negatives,'' in \emph{Advances in Neural Information
  Processing Systems}, 2018, pp. 592--603.

\bibitem{dabkowski2017real}
P.~Dabkowski and Y.~Gal, ``Real time image saliency for black box
  classifiers,'' in \emph{Advances in Neural Information Processing Systems},
  2017, pp. 6967--6976.

\bibitem{he2016deep}
K.~He, X.~Zhang, S.~Ren, and J.~Sun, ``Deep residual learning for image
  recognition,'' in \emph{IEEE Conference on Computer Vision and Pattern
  Recognition}, 2016, pp. 770--778.

\bibitem{baumgartner2017sononet}
C.~F. Baumgartner, K.~Kamnitsas, J.~Matthew, T.~P. Fletcher, S.~Smith, L.~M.
  Koch, B.~Kainz, and D.~Rueckert, ``Sono{N}et: Real-time detection and
  localisation of fetal standard scan planes in freehand ultrasound,''
  \emph{IEEE Transactions on Medical Imaging}, vol.~36, no.~11, pp. 2204--2215,
  2017.

\bibitem{lian2018hierarchical}
C.~Lian, M.~Liu, J.~Zhang, and D.~Shen, ``Hierarchical fully convolutional
  network for joint atrophy localization and {A}lzheimer's disease diagnosis
  using structural {MRI},'' \emph{IEEE Transactions on Pattern Analysis and
  Machine Intelligence}, vol.~42, no.~4, pp. 880--893, 2018.

\bibitem{mirza2014conditional}
M.~Mirza and S.~Osindero, ``Conditional generative adversarial nets,''
  \emph{arXiv preprint arXiv:1411.1784}, 2014.

\bibitem{ronneberger2015u}
O.~Ronneberger, P.~Fischer, and T.~Brox, ``U-{N}et: Convolutional networks for
  biomedical image segmentation,'' in \emph{International Conference on Medical
  Image Computing and Computer-Assisted Intervention}.\hskip 1em plus 0.5em
  minus 0.4em\relax Springer, 2015, pp. 234--241.

\bibitem{zhu2017unpaired}
J.-Y. Zhu, T.~Park, P.~Isola, and A.~A. Efros, ``Unpaired image-to-image
  translation using cycle-consistent adversarial networks,'' in \emph{IEEE
  International Conference on Computer Vision}, 2017, pp. 2223--2232.

\bibitem{goodfellow2016nips}
I.~Goodfellow, ``{NIPS} 2016 tutorial: Generative adversarial networks,''
  \emph{arXiv preprint arXiv:1701.00160}, 2016.

\bibitem{mao2017least}
X.~Mao, Q.~Li, H.~Xie, R.~Y. Lau, Z.~Wang, and S.~Paul~Smolley, ``Least squares
  generative adversarial networks,'' in \emph{IEEE International Conference on
  Computer Vision}, 2017, pp. 2794--2802.

\bibitem{gottlieb1998total}
S.~Gottlieb and C.-W. Shu, ``Total variation diminishing runge-kutta schemes,''
  \emph{Mathematics of Computation}, vol.~67, no. 221, pp. 73--85, 1998.

\bibitem{linsley2018learning}
D.~Linsley, D.~Shiebler, S.~Eberhardt, and T.~Serre, ``Learning what and where
  to attend,'' in \emph{International Conference on Learning Representations},
  2019.

\bibitem{hu2018squeeze}
J.~Hu, L.~Shen, S.~Albanie, G.~Sun, and E.~Wu, ``Squeeze-and-excitation
  networks,'' \emph{IEEE Transactions on Pattern Analysis and Machine
  Intelligence}, vol.~42, no.~8, pp. 2011--2023, 2020.

\bibitem{chen2017deeplab}
L.-C. Chen, G.~Papandreou, I.~Kokkinos, K.~Murphy, and A.~L. Yuille,
  ``{D}eep{L}ab: Semantic image segmentation with deep convolutional nets,
  atrous convolution, and fully connected {CRF}s,'' \emph{IEEE Transactions on
  Pattern Analysis and Machine Intelligence}, vol.~40, no.~4, pp. 834--848,
  2017.

\bibitem{lecun1998mnist}
Y.~LeCun, ``The {MNIST} database of handwritten digits,'' \emph{http://yann.
  lecun. com/exdb/mnist/}, 1998.

\bibitem{kim2018disentangling}
H.~Kim and A.~Mnih, ``Disentangling by factorising,'' in \emph{35th
  International Conference on Machine Learning}, 2018, pp. 2649--2658.

\bibitem{MUELLER2005869}
S.~G. Mueller, M.~W. Weiner, L.~J. Thal, R.~C. Petersen, C.~Jack, W.~Jagust,
  J.~Q. Trojanowski, A.~W. Toga, and L.~Beckett, ``The {A}lzheimer's disease
  neuroimaging initiative,'' \emph{Neuroimaging Clinics of North America},
  vol.~15, no.~4, pp. 869 -- 877, 2005, alzheimer's Disease: 100 Years of
  Progress.

\bibitem{dickerson2001mri}
B.~C. Dickerson, I.~Goncharova, M.~Sullivan, C.~Forchetti, R.~Wilson,
  D.~Bennett, L.~A. Beckett, and L.~deToledo Morrell, ``{MRI}-derived
  entorhinal and hippocampal atrophy in incipient and very mild {A}lzheimer’s
  disease,'' \emph{Neurobiology of Aging}, vol.~22, no.~5, pp. 747--754, 2001.

\bibitem{fan2008spatial}
Y.~Fan, N.~Batmanghelich, C.~M. Clark, C.~Davatzikos, A.~D.~N. Initiative
  \emph{et~al.}, ``Spatial patterns of brain atrophy in {MCI} patients,
  identified via high-dimensional pattern classification, predict subsequent
  cognitive decline,'' \emph{NeuroImage}, vol.~39, no.~4, pp. 1731--1743, 2008.

\bibitem{gerardin2009multidimensional}
E.~Gerardin, G.~Ch{\'e}telat, M.~Chupin, R.~Cuingnet, B.~Desgranges, H.-S. Kim,
  M.~Niethammer, B.~Dubois, S.~Leh{\'e}ricy, L.~Garnero \emph{et~al.},
  ``Multidimensional classification of hippocampal shape features discriminates
  {A}lzheimer's disease and mild cognitive impairment from normal aging,''
  \emph{NeuroImage}, vol.~47, no.~4, pp. 1476--1486, 2009.

\end{thebibliography}

\end{document}